\documentclass{vldb}
\usepackage{graphicx}
\usepackage{balance}  % for  \balance command ON LAST PAGE  (only there!)
\usepackage{algorithm} % For algorithms
\usepackage{stmaryrd}
\usepackage{mathtools}
\usepackage{array}
\usepackage[table]{xcolor}
\usepackage{colortbl}
\usepackage{booktabs,siunitx}
\usepackage[noend]{algpseudocode}
\usepackage{url}

\definecolor{header}{RGB}{130,220,254}
\definecolor{highlight1}{RGB}{255,250,207}
\definecolor{highlight2}{RGB}{146,238,147}
\definecolor{frame}{RGB}{36,147,50}
\definecolor{diff1}{RGB}{254,220,252}
\definecolor{diff2}{RGB}{159,238,255}
\newcolumntype{B}{>{\columncolor[rgb]{0,.6,1}}}

\newcommand*{\colorboxed}{}
\def\colorboxed#1#{%
  \colorboxedAux{#1}%
}
\newcommand*{\colorboxedAux}[3]{%
  \begingroup
    \colorlet{cb@saved}{.}%
    \color#1{#2}%
    \fbox{%
      \color{cb@saved}%
      #3%
    }%
  \endgroup
}

\newtheorem{theorem}{Theorem}[section]

\newtheorem{example}[theorem]{Example}

\newtheorem{definition}[theorem]{Definition}

\begin{document}

% ****************** TITLE ****************************************

\title{Explaining Queries over Web Tables to Non-Experts}

% possible, but not really needed or used for PVLDB:
%\subtitle{[Extended Abstract]
%\titlenote{A full version of this paper is available as\textit{Author's Guide to Preparing ACM SIG Proceedings Using \LaTeX$2_\epsilon$\ and BibTeX} at \texttt{www.acm.org/eaddress.htm}}}

% ****************** AUTHORS **************************************


\numberofauthors{5} %  in this sample file, there are a *total*

\author{
% 1st. author
\alignauthor
Jonathan Berant\\
       \affaddr{Tel Aviv University}\\
       \email{joberant@cs.tau.ac.il}
% 2nd. author
\alignauthor
Daniel Deutch\\
       \affaddr{Tel Aviv University}\\
       \email{danielde@post.tau.ac.il}
% 3rd. author
\alignauthor Amir Globerson\\
       \affaddr{Tel Aviv University}\\
       \email{gamir@mail.tau.ac.il}
\and  % use '\and' if you need 'another row' of author names
% 4th. author
\alignauthor Tova Milo\\
       \affaddr{Tel Aviv University}\\
       \email{milo@cs.tau.ac.il}
% 5th. author
\alignauthor Tomer Wolfson\\
       \affaddr{Tel Aviv University}\\
       \email{tomerwol@mail.tau.ac.il}
}

% Just remember to make sure that the TOTAL number of authors
% is the number that will appear on the first page PLUS the
% number that will appear in the \additionalauthors section.

\maketitle

% Abstract %
\begin{abstract}
Designing a reliable natural language (NL) interface for querying tables has been a longtime goal of researchers in both the data management and natural language processing (NLP) communities. Such an interface receives as input an NL question, translates it into a formal query, executes the query and returns the results. Errors in the translation process are not uncommon, and users typically struggle to understand whether their query has been mapped correctly.  We address this problem by explaining the obtained formal queries to non-expert users. Two methods for query explanations are presented: the first translates queries into NL, while the second method provides a graphic representation of the query cell-based provenance (in its execution on a given table). Our solution augments a state-of-the-art NL interface over web tables, enhancing it in both its training and deployment phase. Experiments, including a user study conducted on Amazon Mechanical Turk, show our solution to improve both the correctness and reliability of an NL interface.
\end{abstract}

%  Introduction %
\section{Introduction}

Natural language interfaces have been gaining significant popularity, enabling ordinary users to write and execute complex queries. One of the prominent paradigms for developing NL interfaces is \emph{semantic parsing}, which is the mapping of NL phrases into a formal language. As Machine Learning techniques are standardly used in semantic parsing, a training set of question-answer pairs is provided alongside a target database \cite{berant2013semantic,pasupat2015wtq,iyer2017learning}. The parser is a parameterized function that is trained by updating its parameters such that questions from the training set are translated into queries that yield the correct answers.

A crucial challenge for using semantic parsers is their reliability. Flawless translation from NL to formal language is an open problem, and even state-of-the-art parsers are not always right. With no explanation of the executed query, users are left wondering if the result is actually correct.
Consider the example in Figure \ref{figure:olympics}, displaying a table of Olympic games and the question \textit{"Greece held its last Olympics in what year?"}. A semantic parser parsing the question generates multiple candidate queries and returns the evaluation result of its top ranked query. The user is only presented with the evaluation result, \textit{2004}. Although the end result is correct, she has no clear indication whether the question was correctly parsed. In fact, the interface might have chosen any candidate query yielding \textit{2004}.
Ensuring the system has executed a correct query (rather than simply returning a correct answer in a particular instance) is essential, as it enables reusing the query as the data evolves over time. For example, a user might wish for a query such as \textit{"The average price of the top 5 stocks on Wall Street"} to be run on a daily basis. Only its correct translation into SQL will consistently return accurate results.

Our approach is to design \textit{provenance-based} \cite{davidson2007provenance, deutch2017nlprov} query explanations that are extensible, domain-independent and immediately understandable by non-expert users. We devise a cell-based provenance model for explaining formal queries over web tables and implement it with our query explanations, (see Figure \ref{figure:olympics}).
We enhance an existing NL interface for querying tables \cite{zhang2017macro} by introducing a novel component featuring our \textit{query explanations}. Following the parsing of an input NL question, our component explains the candidate queries to users, allowing non-experts to choose the one that best fits their intention. The immediate application is to improve the quality of obtained queries at deployment time over simply choosing the parser's top query (without user feedback). Furthermore, we show how query explanations can be used to obtain user feedback which is used to retrain the Machine Learning system, thereby improving its performance.

\begin{figure}[t]
    \centering 
    \begin{minipage}{1\linewidth}
    	\small
        $x$ : "Greece held its last Olympics in what year?" \newline
		$y$ : \textit{\{2004\}} \newline\newline
        \centering \scriptsize
		\begin{tabular}{|B>{\columncolor{highlight1}} c|B>{\columncolor{highlight1}} c|c|} \hline
		\textbf{\color{frame}
		MAX(\color{black}Year\color{frame})} & \textbf{Country} & \textbf{City} \\ \hline
		\colorboxed{frame}{1896} & \colorboxed{frame}{Greece} & Athens \\ \hline
		1900 & France & Paris \\ \hline
		... & ... & ... \\ \hline
		\cellcolor{highlight2}\colorboxed{frame}{2004} & \colorboxed{frame}{Greece} & Athens \\ \hline
		2008 & China & Beijing \\ \hline
		2012 & UK & London \\ \hline
        2016 & Brazil & Rio de Janeiro \\ \hline
		\end{tabular}
        \newline\newline
        \small
		$u$ : \textit{maximum value in column Year where Country is Greece.} 
    %\hspace*{-0.4cm}
    \end{minipage}
    \caption{Querying a table of Olympic games. Above, the user is only shown the final result, $y$. Below, a candidate query is explained by utterance, $u$ and provenance-based highlights.}
    \label{figure:olympics}
    \vspace{-1.5em}
\end{figure}

\paragraph*{Setting} 
We focus on the task of explaining complex queries over web tables to non-expert users. Our solution employs explanations of formal queries to augment an existing state-of-the-art semantic parser \cite{zhang2017macro}. The parser is used as the NL interface, mapping complex NL questions into queries over web tables. We test our solution on real-world data using the \textsc{WikiTableQuestions} benchmark dataset \cite{pasupat2015wtq} which includes over 20,000 complex NL questions formulated by actual users on thousands of extracted web tables.

As our formal query language over tables we use \textit{lambda DCS}, a standard query language in the NLP community \cite{liang2013lambda,berant2013semantic,pasupat2015wtq}. We note that lambda DCS is geared towards queries one would write in a search engine -- such as those in \cite{berant2013semantic,pasupat2015wtq} -- rather than database ones, and as such, its queries receive a single table as their input. However, lambda DCS enables us to formulate highly complex queries supporting operations such as sorting, aggregation and intersection (see Tables \ref{table:wtq_examples} and \ref{table:test_query_comparison} that contain complex NL questions that can be expressed as lambda DCS formulas).

\paragraph*{Contributions} 
We first sought to provide a data provenance model for the lambda DCS query language. In order to align this model with previous works on data provenance we produced a mapping from lambda DCS to SQL. This enabled us to introduce a novel multilevel cell-based provenance model for lambda DCS. Our model was implemented through the use of \textit{provenance-based highlights}. The table highlights serve as query explanations by providing a visual explanation for the underlying query execution (the table in Figure \ref{figure:olympics} shows the highlights for our example query). 
These provenance-based highlights are combined with a more conventional form of query explanation via NL \textit{utterances}. Drawing on previous work \cite{wang2015overnight}, we design an extensible, domain independent context-free grammar that derives NL utterances describing lambda DCS queries. One of our key contributions is showing empirically that combining NL utterances with our provenance-based highlights greatly accelerates user understanding of complex queries.

We demonstrate the effectiveness of our query explanations by further enhancing a semantic parser for querying tables \cite{zhang2017macro} and testing it against a benchmark dataset of thousands of complex NL questions \cite{pasupat2015wtq}. At deployment, users are able to make an informed choice of the query, based on our explanation mechanism. We further leverage the use of explaining queries to users in order to retrain the semantic parser on procured user feedback. Feedback being pairs of NL questions and their correct query. This approach is in line with the \textit{human in the loop} paradigm of users enhancing machine learning systems \cite{holzinger2016interactive,iyer2017learning}.

User studies conducted via Amazon Mechanical Turk (AMT) clearly show our query explanations to be applicable to non-experts. Furthermore, user feedback was beneficial for improving the correctness of generated queries, when used both in deployment and in training time. During deployment, the correctness of obtained queries improved by over 30\% compared to the fully automated process of outputting the top query computed by the baseline parser. 

To summarize, the key contributions of our solution are:
\begin{itemize}
\item \textit{Provenance Model.} We present a novel provenance model for lambda DCS, an expressive query language which has been used in the context of Natural Language Processing question answering.
\item \textit{Query Explanations.} We introduce novel provenance-based highlights for explaining queries over web tables.
\item \textit{Human in the loop.} Our query explanations enable users to impact the parser choices and output. User feedback is crucial in improving the baseline parser when deployed and also for retraining it offline.
\item \textit{User Study.} We show that our query explanations significantly improve the NL interface, while requiring minimal effort from non-expert users. 
\end{itemize}

After reviewing our system (Section \ref{sec:system}) and providing the necessary preliminaries in Section \ref{sec:preliminaries}, we describe our multilevel provenance model in Section \ref{sec:prov_lambda}. Our query explanation methods are presented in Section \ref{sec:explanations}. We discuss the concrete applications of our methods in Section \ref{sec:applications} and measure their contribution via experiments and a user study in Section \ref{sec:experiments}. We discuss related work in Section \ref{sec:related}. In Section \ref{sec:conclusion_future_work} we conclude and point to future work.

\begin{table}[t]
\scriptsize
\centering
\caption{WikiTableQuestions Examples}
\begin{tabular}{p{4.5cm} p{3.5cm}} \hline
  \textbf{Question}  & \textbf{Table Attributes}  \\ \hline
 What was the difference in engine size between Luigi Arcangeli and Louis Chiron? &  No., Driver, Entrant, Constructor, Chassis, Engine\\ \hline
 The US and China are tied in total medal count. Which country has more silver medals, the US or China? & Rank, Nation, Gold, Silver, Bronze, Total \\ \hline
 How long did it take Jeff Lastennet to finish? & Rank, Name, Nationality, Time, Notes, Points \\ \hline
 What's the total number of festivals that occurred in October? & Date, Festival, Location, Awards \\ \hline
 In what position did team Penske finish the first year a Honda engine was used? & Year, Chassis, Engine, Start, Finish, Team\\ \hline
 How many championships were in the \$150,000 category? & Result, Date, Category, Tournament, Surface, Partnering, Opponents, Score\\ \hline
 Which position was recorded the most? & Year, Competition, Venue, Position, Event, Notes\\ \hline
 Which was the only episode to gain a 7 or higher rating? & No., Episode, Air date, Rating, Share, 18-49 (Rating/Share), Viewers (m), Rank (night), Rank (timesolt), Rank (overall)\\ \hline
\end{tabular}
\label{table:wtq_examples}
\vspace{-1.5em}
\end{table}

% System Overview %
\section{System Overview}\label{sec:system}
\begin{figure}
\centering
\includegraphics[trim={11cm 2cm 1cm 1cm},clip, width=7cm,height=5cm]{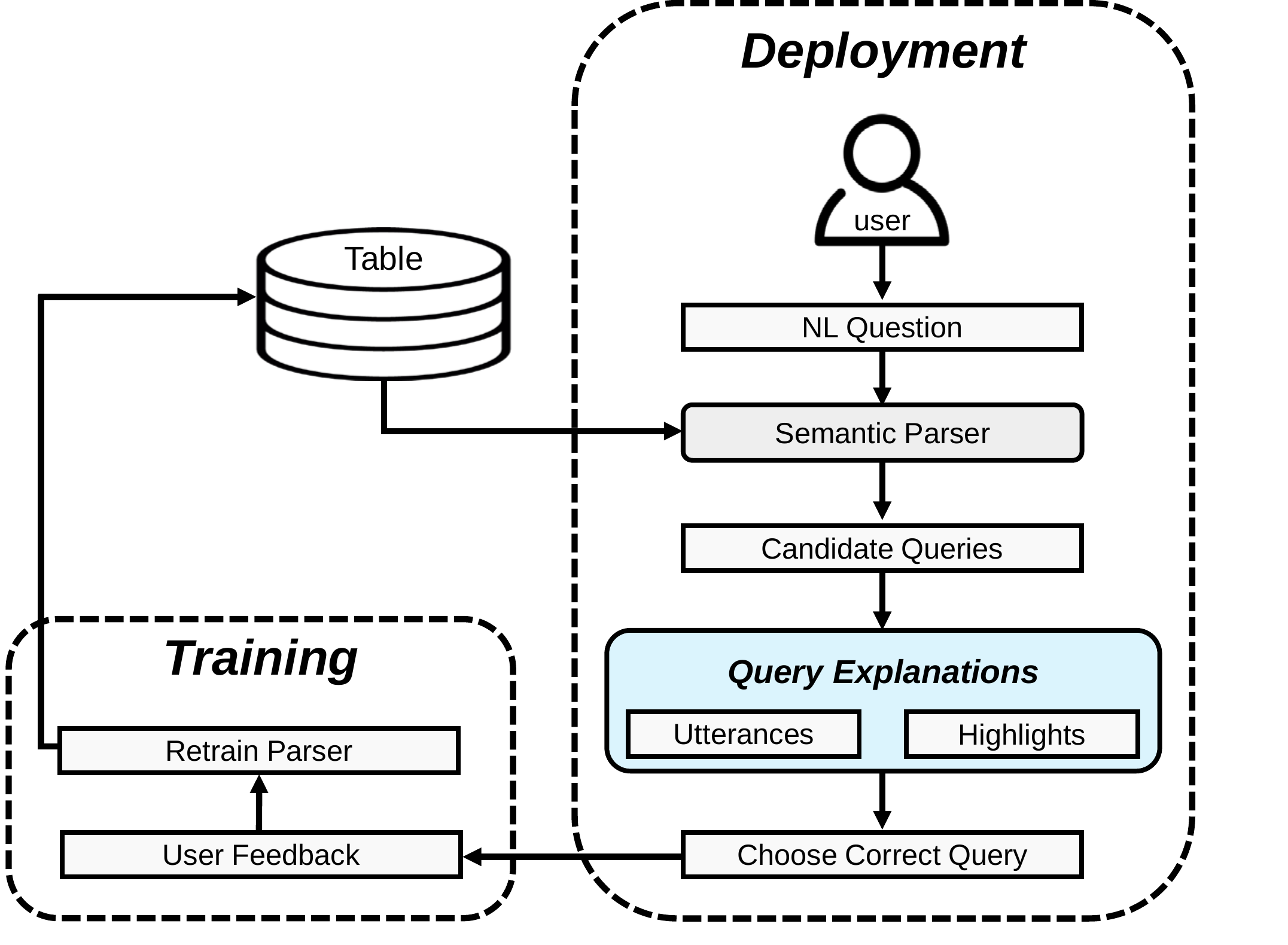}
\caption{System Architecture.}
\label{fig:system}
\vspace{-1.3em}
\end{figure}

We review our system architecture from Figure \ref{fig:system} and describe its general workflow.

\paragraph*{Semantic Parser}
Given an NL question and corresponding table, we use the state-of-the-art parser in \cite{zhang2017macro} to parse the question into a set of candidate lambda DCS queries.
The parser is trained for the task of querying web tables using the \textsc{WikiTableQuestions} dataset \cite{pasupat2015wtq}.

\paragraph*{Query Explanations}
Following the mapping of a question to a set of candidate queries, our interface will generate the relevant query explanations for each query, displaying a detailed NL utterance (Section \ref{sec:utterances}) and highlighting the provenance data using Algorithm \ref{alg:highlights} (Section \ref{sec:prov_highlights}).

\paragraph*{Deployment}
Query explanations are presented to non-technical users to assist in selecting the correct formal-query representing the question.

\paragraph*{Training on Feedback}
User feedback in the form of question-query pairs is used offline in order to retrain the semantic parser.

%  Preliminaries %
\section{Preliminaries}\label{sec:preliminaries}
We begin by formally defining our task of querying tables. Afterwards, we discuss the formal query language and show how lambda DCS queries can be translated directly into SQL. 

% Data Model %
\subsection{Data Model}\label{sec:data_model}
An NL interface for querying tables receives a question $x$ on a table $T$ and outputs a set of values $y$ as the answer (where each value is either the content of a cell, or the result of an aggregate function on cells). 
As discussed in the introduction, we make the assumption that a query concerns a single table.

Following the model presented in \cite{pasupat2015wtq}, all table records are ordered from top to bottom with each record possessing a unique $Index$ (\textit{0, 1, 2, ...}). In addition, every record has a pointer $Prev$ to the record above it. The values of table cells can be either strings, numbers or dates.
While we view the table as a relation, it is common \cite{pasupat2015wtq,zhang2017macro} to describe it as a \textit{knowledge base} (KB) $\mathcal{K}\subset\mathcal{E}\times\mathcal{P}\times\mathcal{E}$ where $\mathcal{E}$ is a set of entities and $\mathcal{P}$ a set of binary properties. The entity set, $\mathcal{E}$ is comprised of all table cells (e.g., $\texttt{Greece}$) and all table records, while $\mathcal{P}$ contains all column headers, serving as binary relations from an entity to the table records it appears in. In the example of Figure \ref{figure:olympics}, column \texttt{Country} is a binary relation such that \texttt{Country.Greece} returns all table records where the value of column Country is Greece (see definition of composition operators below). If the table in Figure \ref{figure:olympics} has $n$ records, the returned records indices will be $\{0, n-4\}$.

% Query Language %
\subsection{Query Language}\label{sec:lambda_dcs}
Following the definition of our data model we introduce our formal query language, \textit{lambda dependency-based compositional semantics} (lambda DCS) \cite{liang2013lambda, berant2013semantic}, which is a language inspired by lambda calculus, that revolves around sets. Lambda DCS was originally designed for building an NL interface over \textsc{Freebase} \cite{bollacker2008freebase}.

\paragraph*{Language Operators}\label{sec:lambda_ops}
Lambda DCS is a highly expressive language, designed to represent complex NL questions involving sorting, aggregation intersection and more. It has been considered a standard language for performing semantic parsing over knowledge bases \cite{liang2013lambda,berant2013semantic,pasupat2015wtq,zhang2017macro}.
A lambda DCS formula is executed against a target table and returns either a set of values (string, number or date) or a set of table records. We describe here a simplified version of lambda DCS that will be sufficient for understanding the examples presented in this paper. For a full description of lambda DCS, the reader should refer to \cite{liang2013lambda}. The basic constructs of lambda DCS are as follows:
\begin{itemize}
  \item \textbf{Unary:} a set of values. The simplest type of unary in a table is a table cell, e.g., \texttt{Greece}, which denotes the set of cells containing the entity 'Greece'. 
  \item \textbf{Binary:} A binary relation describes a relation between sets of objects. The simplest type of a binary relation is a table column $C \in \mathcal{P}$, mapping table entities to the records where they appear, e.g., \texttt{Country}.
  \item \textbf{Join:} For a binary relation $C$ and unary relation $v$, $C.v$ operates as a selection and projection. $\texttt{Country.Greece}$ denotes all table records where the value of column Country is Greece.
\item \textbf{Prev:} Given records $\{r_{i_{0}},r_{i_{1}},...,r_{i_{n}}\}$ the $Prev$ operator will return the set of preceding table records, $Prev[\{r_{i_{0}},r_{i_{1}},...,r_{i_{n}}\}] = \{r_{i_{0}-1},r_{i_{1}-1},...,r_{i_{n}-1}\}$.  
  \item \textbf{Reverse:} Given a binary relation $b$ from $s$ to $t$, there is a reversed binary relation \textbf{R}[$b$] from $t$ to $s$. E.g., for a column binary relation $C$ from table values to their records, \textbf{R}[$C$] is a relation from records to values. \texttt{R[Year].Country.Greece} takes all the record indices of \texttt{Country.Greece} and returns the values of column Year in these records. Similarly, \texttt{R[Prev]} denotes a relation from a set of records, to the set of following (reverse of previous) table records. 
  \item \textbf{Intersection:} Intersection of sets. E.g., the set of records where Country is Greece and also where Year is 2004, \texttt{Country.Greece $\sqcap$ Year.2004}.
  \item \textbf{Union:} Union of sets. E.g., records where the value of column Country is Greece or China, \texttt{Country.Greece $\sqcup$ Country.China}.
  \item \textbf{Aggregation:} Aggregate functions \texttt{min, max, avg, sum, count} that take a unary and return a unary with one number. E.g., $\texttt{count(City.Athens)}$ returns the number of records where the value of City is Athens.
  \item \textbf{Superlatives:} \texttt{argmax, argmin}. For unary $u$ and binary $b$, $\texttt{argmax(u,b)}$ is the set of all values $x \in \texttt{argmax(u,b)}$.
\end{itemize}

\paragraph*{Compositional Operators} In this paper we use a group of predefined operators specifically designed for the task of querying tables \cite{pasupat2015wtq}. The language operators are \textit{compositional} in nature, allowing the semantic parser to compose several sub-formulas into a single formula representing complex query operations.

\begin{example}
Consider the following lambda DCS query on the table from Figure \ref{figure:olympics}, 
$$
\vspace{-0.25em}
\texttt{R[City].argmin(Record, Year)} 
\vspace{-0.07em}
$$
it returns values of column City (binary) appearing in records (Record unary) that have the lowest value in column Year.
\end{example} 

\paragraph*{Mapping to SQL}\label{sec:lambda_sql}
To position our work in the context of relational queries we show lambda DCS to be an expressive fragment of SQL. The translation into SQL proves useful when introducing our provenance model by aligning our model with previous work \cite{glavic2013using,deutch2017nlprov}.
Table \ref{table:lambdaDCS} (presented at the end of the paper) describes all lambda DCS operators with their corresponding translation into SQL.

\begin{example}
Returning to the lambda DCS query from the previous example, it can be easily translated to SQL as,
\vspace{2mm}\newline
\texttt{
\hspace*{0.35cm}SELECT City FROM T\newline
\hspace*{0.35cm}WHERE Index IN (\newline
\hspace*{1.2cm}SELECT Index FROM T \newline
\hspace*{1.2cm}WHERE Year = ( SELECT MIN(Year) FROM T ) );
}\vspace{2mm}
\newline
where \texttt{Index} denotes the attribute of record indices in table $T$. The query first computes the set of record indices containing the minimum value in column Year, which in our running example table is \{0\}. It then returns the values of column City in these records, which is \textit{Athens} as it is the value of column City at record 0.  
\end{example}

\begin{table*}[t]
\scriptsize 
\centering
\caption{Example provenance for several query operators}
\begin{tabular}
{|p{1.5cm}|p{2.3cm}|p{3.5cm}|p{3.2cm}|p{5.5cm}|}\hline
\textbf{Operator} & \textbf{Query (lambda DCS)} & \textbf{Example} & \textbf{Translation} & \textbf{Provenance}\\ \hline
Column Values & \texttt{\textbf{R}[C].records} & \texttt{\textbf{R}[Year].City.Athens} & "value of column Year where City is Athens" & $P_O(Q) = \{c \mid c \in C \; \land \; record(c).Index \in records.Indices\} $ \newline $P_E(Q) = P_O(Q) \cup P_E(records) $ \newline $P_C(Q) = \{c \mid c \in C\} \cup P_C(records)$\\ \hline
Aggregation on Values & \texttt{aggr(vals)} \newline \textcolor{gray}{aggr $\in$ \{count, max, min, sum, avg\}} & \texttt{sum(R[Year].City.Athens)} & "the sum of values in column Year where City is Athens" & $P_O(Q) = P_O(vals) \cup \{AGGR\} $ \newline $P_E(Q) = P_O(Q)$ \newline $P_C(Q) = \{c \mid c \in C \}$\\ \hline
Difference of Values & \texttt{sub(\textbf{R}[C1].C2.v, \textbf{R}[C1].C2.u)} & \texttt{sub(\textbf{R}[Year].City.London, \textbf{R}[Year].City.Beijing)} & "the difference in column Year between London and Beijing" & $P_O(Q) = P_O(R[C_1].C_2.v) \cup P_O(R[C_1].C_2.u)$ \newline $P_E(Q) = P_E(R[C_1].C_2.v) \cup P_E(R[C_1].C_2.u)$ \newline $P_C(Q) = \{c \mid c \in C_1 \; \lor \; c \in C_2\}$\\ \hline
Intersection of Records & \texttt{records\textsubscript{1} $\sqcap$ records\textsubscript{2}} & \texttt{City.London $\sqcap$ Country.UK} & "rows where value of City is London and also where value of Country is UK" & $P_O(Q) = P_O(records_1) \cap P_O(records_2) $ \newline $P_E(Q) = P_E(records_1) \cup P_E(records_2)$ \newline $P_C(Q) = P_C(records_1) \cup P_C(records_2)$\\ \hline
\end{tabular}
\label{table:lambdaDCS_provenance}
\vspace{-1.5em}
\end{table*}

% Provenance %
\section{Provenance}\label{sec:prov_lambda}
The tracking and presentation of provenance data has been extensively studied in the context of relational queries \cite{glavic2013using, deutch2017nlprov}. In addition to explaining query results \cite{deutch2017nlprov}, we can use provenance information for explaining the query execution on a given web table. We design a model for \textit{multilevel cell-based provenance} over tables, with three levels of granularity. The model enables us to distinguish between different types of table cells involved in the execution process. This categorization of provenance cells serves as a form of query explanation that is later implemented in our provenance-based highlights (Section \ref{sec:prov_highlights}).

\subsection{Model Definitions}
Given query $Q$ and table $T$, the execution result, denoted by $Q(T)$, is either a collection of table cells, or a numeric result of an aggregate or arithmetic operation.

We define $\mathcal{Q}$ to be the infinite domain of possible queries over $T$, $records(T)$ to be the set of table records, $cells(T)$ to be the set of table cells and denote by $aggrs$ the set of aggregate functions, \textit{\{min, max, avg, count, sum\}}. 

Our cell-based provenance takes as input a query and its corresponding table and returns the set of cells and aggregate functions involved in the query execution. The model distinguishes between three types of provenance cells. There are the cells returned as the query output $Q(T)$, cells that are examined during the execution, and also the cells in columns that are projected or aggregated on by the query. We formally define the following three cell-based provenance functions.

\begin{definition}
Let $Q$ be a formal query and $T$ its corresponding table. We define three \textit{cell-based provenance functions}, $P_O(Q, T), P_E(Q, T), P_C(Q, T)$. Given $Q,T$ the functions output a set of table cells and aggregate functions. 
\begin{displaymath}
P_*(Q, T): \mathcal{Q} \times records(T) \longrightarrow 2^{cells(T)} \cup 2^{aggrs}
\end{displaymath}

We use $OP$ to denote an aggregate function or arithmetic operation on tables cells. Given the compositional nature of the lambda DCS query language, we define $Q_{SUB}$ as the set of all sub-queries composing $Q$. We have used $C \in Q$ to denote the table columns that are either projected by the query, or that are aggregated on by it.

\begin{equation} \label{eq:provoutput}
  \begin{split}
  P_O(Q,T) \coloneqq 
  \begin{cases}
    P_O(Q',T) \cup \{OP\}, & Q \equiv OP(Q').\\
    \{c_1,...,c_n \in Q(T)\}, & \text{otherwise}.
  \end{cases}
  \end{split}
  \end{equation}
\begin{equation}
  P_E(Q,T) \coloneqq \bigcup_{Q' \in Q_{SUB}}P_O(Q',T)
  \end{equation}
\begin{equation}
   P_C(Q,T) \coloneqq \{c \in C \mid C \in Q\}
  \end{equation}
\end{definition}

Function $P_O(Q,T)$ returns all cells output by $Q(T)$ or, if $Q(T)$ is the result of an arithmetic or aggregate operation, returns all table cells involved in that operation in addition to the aggregate function itself.
$P_E(Q,T)$ returns cells and aggregate functions used during the query execution.
$P_C(Q,T)$ returns all table cells in columns that are either projected or aggregated on by $Q$.
These cell-based provenance functions have a hierarchical relation, where the cells output by each function are a subset of those output by the following function. Therefore, the three provenance sets constitute an \textit{ordered chain}, where $P_O(Q,T) \subseteq P_E(Q,T) \subseteq P_C(Q,T)$.

Having described our three levels of cell-based provenance, we combine them into a single \textit{multilevel cell-based model} for querying tables.

\begin{definition}
Given formal query $Q$ and table $T$, the \textit{multilevel cell-based provenance} of $Q$ executed on $T$ is a function, 
\begin{displaymath}
Prov(Q, T): \mathcal{Q} \times records(T) \longrightarrow \{2^{cells(T)} \cup 2^{aggrs}\}^3.
\end{displaymath}
Returning the provenance chain,
\begin{displaymath}
Prov(Q, T) \coloneqq (P_O(Q,T), P_E(Q,T), P_C(Q,T)).
\end{displaymath}

\end{definition}

\subsection{Query Operators} 
Using our model, we describe the multilevel cell-based provenance of several lambda DCS operator in Table \ref{table:lambdaDCS_provenance}. Provenance descriptions of all lambda DCS operators are provided in Table \ref{table:lambdaDCS} (at the end of the paper). 
For simplicity, we omit the table parameter $T$ from provenance expressions, writing $P_*(Q)$ instead of $P_*(Q,T)$. We also denote both cells and aggregate functions as belonging to the same set.

We use $c$ to denote a table cell with value $T[c]$, while denoting specific cell values by $u, v$. Each cell $c$ belongs to a table record, $record(c)$ with a unique index, $record(c).Index$ (Section \ref{sec:data_model}). 
We distinguish between two types of lambda DCS formulas: formulas returning values are denoted by $vals$ while those returning table records by $records$.

\begin{example}
We explain the provenance of the following lambda DCS query, 
$$
\texttt{R[Year].City.Athens} 
$$
It returns the values of column Year in records where column City is Athens, thus $P_O(Q,T)$ will return all cells containing these values.
\begin{multline*}
P_O(R[Year].City.Athens, T) = \{c \mid c \in Year \; \land \\ record(c).Index \in City.Athens.Indices\}
\end{multline*}
The cells involved in the \textbf{execution} of $Q$ include the output cells $P_O(Q,T)$ in addition to the provenance of the sub-formula \texttt{City.Athens}, defined as all cells of column City with value Athens.
\begin{multline*}
P_E(R[Year].City.Athens, T) = \\ P_O(R[Year].City.Athens, T) \; \cup \; P_O(City.Athens, T) \\
\end{multline*}
Where, 
\begin{multline*}
P_O(City.Athens, T) \;=\; \{c \mid c \in City \; \land \; T[c]=Athens\}.
\end{multline*}
The provenance of the \textbf{columns} of $Q$ is simply all cells appearing in columns Year and City.
\begin{multline*}
P_C(R[Year].City.Athens, T) = \{c \mid c \in Year \; \lor \; c \in City\}
\end{multline*}
The provenance rules used in the examples regard the lambda DCS operators of "column records" and of "column values". The definition of the relevant provenance rules are described in the first two rows of Table \ref{table:lambdaDCS}.
\end{example}

% Explaining Queries %
\section{Explaining Queries}\label{sec:explanations}
To allow users to understand formal queries we must provide them with effective explanations.
We describe the two methods of our system for explaining its generated queries to non-experts.
Our first method translates formal queries into NL, deriving a detailed utterance representing the query. The second method implements the multilevel provenance model introduced in Section \ref{sec:prov_lambda}. For each provenance function ($P_O, P_E, P_C$) we uniquely highlight its cells, creating a visual explanation of the query execution.

% Query to Utterance %
\subsection{Query to Utterance}\label{sec:utterances}
Given a formal query in lambda DCS we provide a domain independent method for converting it into a detailed NL utterance.
Drawing on the work in \cite{wang2015overnight} we use a similar technique of deriving an NL utterance alongside the formal query. We introduce new NL templates describing complex lambda DCS operations for querying tables. 

\begin{figure}
\centering

\includegraphics[trim={11.3cm 0cm 0cm 6cm},clip, width=9cm,height=5cm]{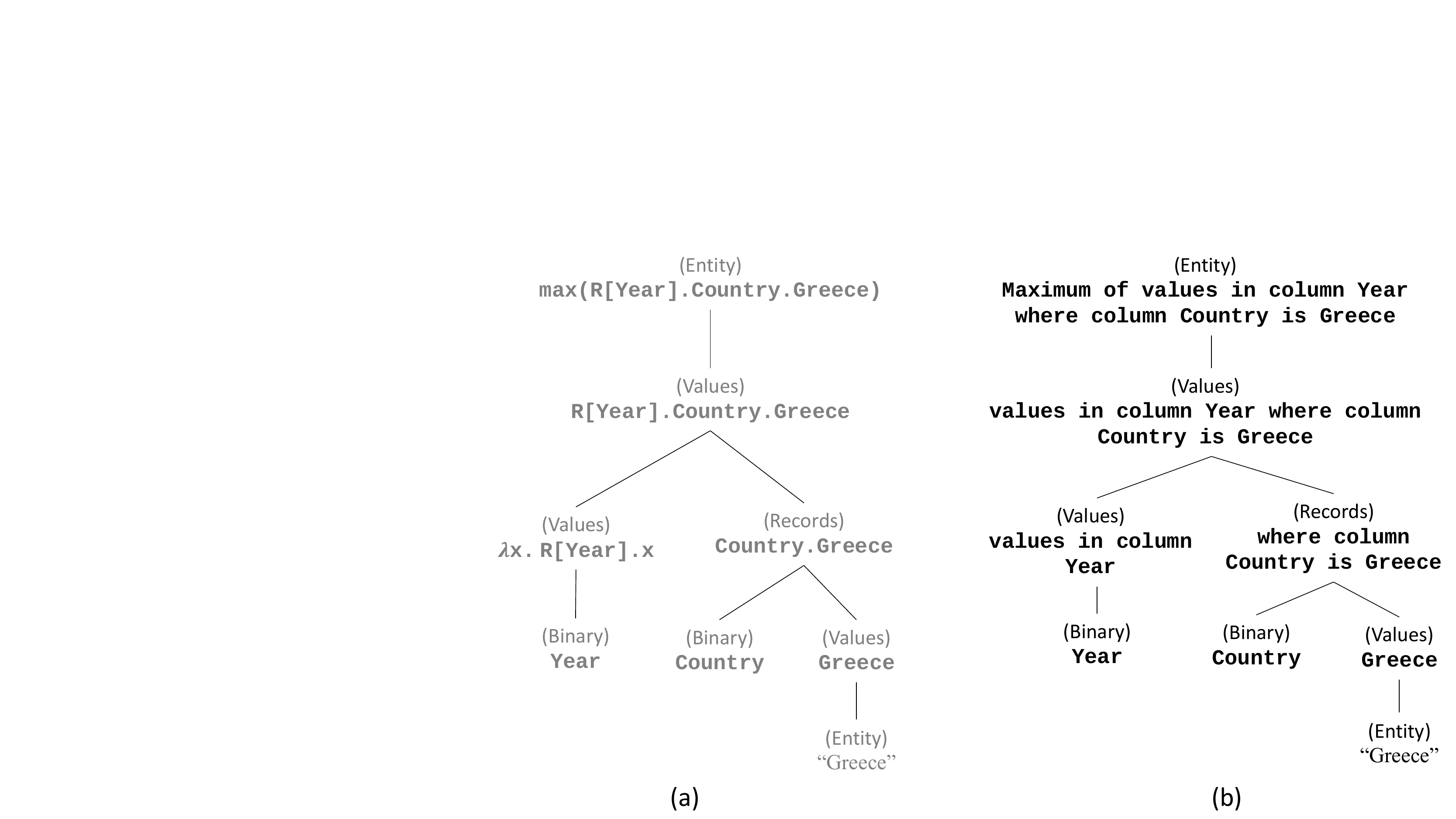}
\caption{(a) The parser's derivation tree of the formal query. (b) Our derived NL utterance explaining the query. Derivations are composed bottom-up.}
\label{fig:derivation}
\vspace{-1.5em}
\end{figure}

\begin{example}
The lambda DCS query,
$$\texttt{\textbf{R}[Year].Country.Greece}$$
is mapped to the utterance, \textit{"value in column Year where column Country is Greece"}. If we compose it with an aggregate function,
$$\texttt{max(\textbf{R}[Year].Country.Greece)}$$
its respective utterance will be composed as well, being \textit{"maximum of values in column Year where column Country is Greece"}. The full derivation trees are presented in Figure \ref{fig:derivation}, where the original query parse tree is shown on the left, while our derived NL explanation is presented on the right.
\end{example}

We implement \textit{query to utterance} as part of the semantic parser of our interface (Section \ref{sec:semanticparsing}). The actual parsing of questions into formal queries is achieved using a context-free grammar (CFG). As shown in Figure \ref{fig:derivation}, formal queries are derived recursively by repeatedly applying the grammar deduction rules. Using the CYK \cite{kasami1966efficient} algorithm, the semantic parser returns derivation trees that maximize its objective (Section \ref{sec:semanticparsing}).
To generate an NL utterance for any formal query, we change the right-hand-side of each grammar rule to be a sequence of both non-terminals and NL phrases. 
For example, grammar rule: (\textit{"maximum of"} Values $\rightarrow$ Entity) where Values, Entity and \textit{"maximum of"} are its non-terminals and NL phrase respectively. Table \ref{table:utterances} describes the rules of the CFG augmented with our NL utterances. 
At the end of the derivation, the full query utterance can be read as the yield of the parse tree.

To utilize utterances as query explanations, we design them to be as clear and understandable as possible, albeit having a somewhat clumsy syntax. The references to table \textit{columns, rows} as part of the NL utterance helps to clarify the actual semantics of the query to the non-expert users.

As the utterances are descriptions of formal queries, reading the utterance of each candidate query to determine its correctness might take some time. As user work-time is expensive, explanation methods that allow to quickly target correct results are necessary. We enhance utterances by employing provenance-based explanations, used for quickly identifying correct queries.

\begin{table*}[tp]
\centering
\scriptsize
\caption{Parser grammar combined with utterances}
\begin{tabular}{|p{9.2cm}|p{8.6cm}|}\hline
\textbf{Rule} & \textbf{Example Utterance}\\ \hline
Entity $\rightarrow$ Values & \textit{Athens.}\\ \hline
\textit{"is at most"} Entity $\rightarrow$ Values & \textit{is at most 17.}\\ \hline
\textit{"rows where value in column"} Binary \textit{"is"} Values $ \rightarrow$ Records & \textit{rows where value in column City is Athens or London.}\\ \hline
\textit{"values in column"} Binary \textit{"in rows"} Records $\rightarrow$ Values & \textit{values of column Year in rows where value of column City is Athens.}\\ \hline
\textit{"right above"} Records $\rightarrow$ Records & \textit{right above rows where value of column City is Athens.}\\ \hline
\textit{"the number of"} Records $\rightarrow$ Entity & \textit{the number of rows where value of column City is Athens.}\\ \hline
\textit{"maximum of"} Values $\rightarrow$ Entity & \textit{maximum of values in column Year in rows where value of column City is Athens.}\\ \hline
\textit{"difference in value of column"} ValueFunc  Values \textit{"and"} Values $\rightarrow$ Values \newline Binary \textit{"between rows where"} Binary \textit{"is"} $\rightarrow$ ValueFunc & \textit{difference in values of column Year between rows where values of column City is London and Beijing.}\\ \hline
\textit{"in column"} Binary \textit{"what is the difference between rows with value"} Entity \textit{"and rows with value"} Entity $\rightarrow$ Values & \textit{in column City, what is the difference between rows with value Athens and rows with value London.}\\ \hline
Entity \textit{"or"} Entity $\rightarrow$ Values & \textit{China or Greece.}\\ \hline
Records \textit{"and also"} Records $\rightarrow$ Records & \textit{rows where value of column City is London and also where value of column Country is UK.}\\ \hline
Records \textit{"that have the highest value in column"} Binary $\rightarrow$ Records & \textit{rows that have the highest value in column Year.}\\ \hline
\textit{"where it is the last row"} Records $\rightarrow$ Records & \textit{where it is the last row in rows where value of column City is Athens.}\\ \hline
\textit{"the value of"} Values \textit{"that appears the most in column"} Binary $\rightarrow$ Values & \textit{the value of Athens or London that appears the most in column City.}\\ \hline
\textit{"between"} Values \textit{"who has the highest value of column"} Binary $\rightarrow$ Values & \textit{between London or Beijing who has the highest value of column Year.}\\ \hline

\end{tabular}
\label{table:utterances}
\vspace{-1.5em}
\end{table*}

% Provenance to Highlights %
\subsection{Provenance to Highlights}\label{sec:prov_highlights}
The understanding of a table query can be achieved by examining the cells on which it is executed. We explain a query by highlighting its multilevel cell-based provenance (Section \ref{sec:prov_lambda}). 

Using our provenance model, we define a procedure that takes a query as input and returns all cells involved in its execution on the corresponding table. These cells are then highlighted in the table, illustrating the query execution.
Given a query $Q$ and table $T$, the $\textsc{Highlight}(Q, T, output)$ procedure divides cells into four types, based on their multilevel provenance functions. To help illustrate the query, each type of its provenance cells is highlighted differently: \textit{Colored cells} are equivalent to $P_O(Q,T)$ and are the cells returned by $Q$ as output, or used to compute the final output. \textit{Framed cells} are equivalent to $P_E(Q,T)$ and are the cells and aggregate functions used during query execution. \textit{Lit cells} are equivalent to $P_C(Q,T)$, and are the cells of columns projected by the query. All other cells are unrelated to the query, hence no highlights are applied to them.

\begin{example}
Consider the lambda DCS query,
$$
\small
\texttt{sub(\textbf{R}[Total].Nation.Fiji, \textbf{R}[Total].Nation.Tonga)}.
$$
The utterance of this query is, \textit{"difference in column Total between rows where Nation is Fiji and Tonga"}.
Figure \ref{fig:highlight_diff_val} displays the highlights generated for this query, lighting all of the query's columns, framing its provenance cells and coloring the cells that comprise its output.
In this example, all cells in columns \textit{Nation} and \textit{Total} are lit. The cells \textit{Fiji} and \textit{Tonga} are part of $P_E(Q,T)$ and are therefore framed. The cells in $P_O(Q,T)$, containing \textit{130} and \textit{20}, are colored as they contain the values used to compute the final result.
\end{example}

To highlight a query over the input table we call the procedure $\textsc{Highlight}(Q, T, output)$ with $output = true$. We describe our implementation in Algorithm \ref{alg:highlights}. It is a recursive procedure which leverages the compositional nature of lambda DCS formulas. It decomposes the query $Q$ into its set of sub-formulas $Q_{SUB}$, recursively computing the multilevel provenance. When reaching an atomic formula the algorithm will execute it and return its output. Cells returned by a sub-formula are both lit and framed, being part of $P_C(Q,T)$ and $P_E(Q,T)$. Finally, all of the cells in $P_O(Q,T)$ (Equation \ref{eq:provoutput}) are colored. 

Examples of provenance-based highlights are provided for several lambda DCS operators in Figures \ref{fig:highlight_comparison}-\ref{fig:highlight_diff_val}. We display highlight examples for all lambda DCS operators in Figures \ref{tbl:highlight_join} - \ref{tbl:highlight_superlative_occur} (at the end of the paper).
\begin{algorithm}
\small
\caption{Highlighting query cell-based provenance}\label{euclid}
\begin{algorithmic}[1]
\Procedure{Highlight}{$Q$, $T$, $output$}
\State $P_O, P_E, P_C \gets \varnothing$ \Comment{provenance sets}
\State $P_O \gets P_O \cup \{Q(T)\}$
\If{$AGGR \in Q$} \Comment{aggregate function}
\State $\Call{MarkColumnHeader}{AGGR, Q, T}$
\EndIf
\If{$Q$ is atomic} 
\State $P_E \gets P_O$
\Else{ 
\State $Q_{SUB} \gets \Call{Decompose}{Q}$
\For{$Q' \in Q_{SUB}$}
\State $P'_O,P'_E,P'_C \gets \Call{Highlight}{Q', T, false}$
\State $P_E \gets P_E \cup P'_E $
\EndFor
}\EndIf
\For{$C \in Q$}
\For{$c \in C$}
\State $P_C \gets P_C \cup \{c\}$
\EndFor
\EndFor
\If{$output = True$} 
\State $\Call{LitCells}{P_C, T}$; $\Call{FrameCells}{P_E, T}$
\State $\Call{ColorCells}{P_O, T}$
\EndIf
\State\Return $(P_O, P_E, P_C)$
\EndProcedure
\end{algorithmic}
\label{alg:highlights}
\end{algorithm}

We note that different queries may possess identical provenance-based highlights. Consider Figure \ref{fig:highlight_comparison} and the following query utterances,
\begin{enumerate}
  \item \textit{"values in column Games that are more than 4."}
  \item \textit{"values in column Games that are at least 5 and also less than 17."}
\end{enumerate}
The highlights displayed on Figure \ref{fig:highlight_comparison} will be the same for both of the above queries. In such cases the user should refer to the NL utterances of the queries in order to distinguish between them. Thus our query explanation methods are complementary, with the provenance-based highlights providing quick visual feedback while the NL utterances serve as detailed descriptions.

\begin{figure*}[t]
    \centering \scriptsize
      
    \begin{minipage}{.3\linewidth}
        \centering\tiny
         \begin{tabular}{|c | c |B>{\columncolor{highlight1}} c | c | c |}
            \hline \textbf{Name} & \textbf{Position} & \textbf{Games} &  ...\\
            \hline Erich Burgener & GK & 3 &  ...\\
            \hline Charly In-Albon & DF & 4 &  ... \\
            \hline Andy Egli & DF & \cellcolor{highlight2}\colorboxed{frame}{6} &  ...\\
            \hline Marcel Koller & DF & 2 &  ...\\
            \hline Heinz Hermann & MF & \cellcolor{highlight2}\colorboxed{frame}{6} &  ...\\
            \hline Lucien Favre & MF & \cellcolor{highlight2}\colorboxed{frame}{5} &  ...\\
            \hline ... & ... & ... & ...\\
            \hline
    %\hspace*{-0.4cm}
        \end{tabular}
        \caption{Comparison}\label{fig:highlight_comparison}
        \small
        \textit{rows where values of column Games are more than 4.\newline}
    \end{minipage}
    \begin{minipage}{.3\linewidth}
        \centering\tiny
        \begin{tabular}{|B>{\columncolor{highlight1}} c | c |B>{\columncolor{highlight1}} c | c | c | c |}
            \hline \textbf{Year} & \textbf{Country} & \textbf{City} & ...\\ \hline
			1896 & Greece & Athens & ...\\ \hline
			1900 & France & Paris & ...\\ \hline
			... & ... & ... & ...\\ \hline
			2004 & Greece & Athens & ...\\ \hline
			\colorboxed{frame}{2008} & China & \colorboxed{frame}{Beijing} & ...\\ \hline
			\cellcolor{highlight2}\colorboxed{frame}{2012} & UK & \colorboxed{frame}{London} & ...\\ \hline
            2016 & Brazil & Rio de Janeiro & ...\\ \hline
    %\hspace*{-0.4cm}
        \end{tabular}
        \caption{Superlative (values)}\label{fig:highlight_superlative_val}
        \small
        \textit{between London or Beijing who has the highest value of column Year.\newline}
    \end{minipage}
    \begin{minipage}{.3\linewidth}
        \centering\tiny
        \begin{tabular}{|c |B>{\columncolor{highlight1}} c | c | c |B>{\columncolor{highlight1}}c |}
            \hline \textbf{Rank} & \textbf{Nation} & \textbf{Gold} & ... & \textbf{Total} \\ \hline
			1 & New Caledonia & 120 & ... & 288\\ \hline
            2 & Tahiti & 60 & ... &  144\\ \hline
            3 & Papua New Guinea & 48 & ... &  121\\ \hline
            4 & \colorboxed{frame}{Fiji} & 33 & ... & \cellcolor{highlight2}\colorboxed{frame}{130} \\ \hline
            5 & Samoa & 22 & ... &  73\\ \hline
            6 & \colorboxed{frame}{Tonga} & 4 &  ... &  \cellcolor{highlight2}\colorboxed{frame}{20} \\ \hline
            ... & ... & ... & ... & ...\\ \hline
    %\hspace*{-0.4cm}
        \end{tabular}
        \caption{Difference (values)}\label{fig:highlight_diff_val}
        \small
        \textit{difference in column Total between Fiji and Tonga.\newline}
    \end{minipage}
\vspace{-1.5em}
\end{figure*}
\newpage

% Scaling to Large Tables %
\subsection{Scaling to Large Tables}\label{sec:scaling}
We elaborate on how our query explanations can be easily extended to tables with numerous records. Given the nature of the NL utterances, this form of explanation is independent of a table's given size. The utterance will still provide an informed explanation of the query regardless of the table size or its present relations.

When employing our provenance-based highlights to large tables it might seem intractable to display them to the user. However, the highlights are meant to explain the candidate query itself, and not the final answer returned by it. Thus we can precisely indicate to the user what are the semantics of the query by employing highlights to a subsample of the table.

An intuitive solution can be used to achieve a succinct sample. First we use Algorithm \ref{alg:highlights} to compute the cell-based provenance sets $(P_O(Q,T), P_E(Q,T), P_C(Q,T))$ and to mark the aggregation operators on relevant table headers. We can then map each provenance cell to its relevant record (table row), enabling us to build corresponding record sets, $R_O(Q,T), R_E(Q,T), R_C(Q,T)$. To illustrate the query highlights we sample one record from each of the three sets: $R_O(Q,T)$, $R_E(Q,T) \smallsetminus R_O(Q,T)$ and $R_C(Q,T) \smallsetminus R_E(Q,T)$. In the special case of a query containing arithmetic difference (Figure \ref{fig:highlight_diff_val}), we select two records from $R_O(Q,T)$, one for each subtracted value. Sampled records are ordered according to their order in the original table. The example in Figure \ref{figure:sampling_highlights} contains three table rows selected from a large web table \cite{googleBigQueryPublicData}. 

\begin{figure}[t]
    \centering 
    \begin{minipage}{1\linewidth}
    	\small
        "What was the highest growth rate of Madagascar in the 1980s?" \newline\newline
        \centering \scriptsize
		\begin{tabular}{|c|B>{\columncolor{highlight1}} c|B>{\columncolor{highlight1}} c|c|B>{\columncolor{highlight1}} c|} \hline
		\textbf{Row} & \textbf{Country} & \textbf{Year} & ... & \textbf{\color{frame} MAX(\color{black}Growth Rate\color{frame})} \\ \hline
        14266 & \colorboxed{frame}{Madagascar} & \colorboxed{frame}{1986} & ... & \cellcolor{highlight2}\colorboxed{frame}{2.945} \\ \hline
        14270 & \colorboxed{frame}{Madagascar} & \colorboxed{frame}{1983} & ... & \colorboxed{frame}{2.877}	 \\ \hline
        14454 & Burkina Faso & 2011 & ... & 3.085 \\ \hline
		\end{tabular} 
    %\hspace*{-0.4cm}
    \end{minipage}
    \caption{Scaling highlights to a large table by selecting three table rows.}
    \label{figure:sampling_highlights}
    \vspace{-1.5em}
\end{figure}

% Applications %
\section{Concrete Applications}\label{sec:applications}
So far we have described our methods for query explanations (Sections \ref{sec:utterances}, \ref{sec:prov_highlights}) and we now harness these methods to enhance an existing NL interface for querying tables.

\paragraph*{Deployment} When deployed, our interface is given a table and a corresponding user question. It parses the question, generating a set of candidate queries ranked by their likelihood of being correct. We display the top-k candidates to users using our explanations (utterances and highlights). The choice of $k$ is discussed in Section \ref{sec:experiments_deployment}. Through the query explanations users can identify which queries are correct translations of the question, and which should be discarded. If no correct query was generated among the parser's top-k candidates, the user should mark \textit{None}. This allows users to choose which queries are to be executed, substituting the system-selected query when necessary and thereby improving its overall correctness.

\paragraph*{Training on Feedback} User feedback is also used to enhance the system correctness through training. 
The mapping from NL to formal queries is learned by the semantic parser of \cite{zhang2017macro}, trained on a large scale dataset for querying web tables \cite{pasupat2015wtq}. We improve the semantic parser by retraining it on pairs of questions and formal queries, marked as correct translations by users.
While it is known that training a semantic parser on question-query pairs improves its performance \cite{artzi2013weakly}, up until now the only way to achieve this was by relying on expert annotators. We are the first to illicit such annotations without any reliance on experts \cite{artzi2013weakly, yih2016value,iyer2017learning}.

% Implementation %
\subsection{Implementation}
We return to our system architecture from Figure \ref{fig:system}.
Presented with an NL question and corresponding table, our interface parses the question into lambda DCS queries using the state-of-the-art parser in \cite{zhang2017macro}. The parser is trained for the task of querying web tables using the \textsc{WikiTableQuestions} dataset \cite{pasupat2015wtq}.

Following the mapping of a question to a set of candidate queries, our interface will generate relevant query explanations for each of the queries, displaying a detailed NL utterance and highlighting the provenance data. The explanations are presented to non-technical users to assist in selecting the correct formal-query representing the question.

User feedback in the form of question-query pairs is also used offline in order to retrain the semantic parser.

We briefly describe the benchmark dataset used in our framework and its relation to the task of querying web tables.

% WikiTable Questions %
\paragraph*{WikiTableQuestions Dataset}\label{sec:wtq}
\textsc{WikiTableQuestions} \cite{pasupat2015wtq} is a question answering dataset over semi-structured tables. It is comprised of question-answer pairs on HTML tables, and was constructed by selecting data tables from Wikipedia that contained \textit{at least} 8 rows and 5 columns. Amazon Mechanical Turk workers were then tasked with writing trivia questions about each table. In contrast to common NLIDB benchmarks \cite{iyer2017learning,berant2013semantic,jagadish2014nalir}, \textsc{WikiTableQuestions} contains 22,033 questions and is an order of magnitude larger than previous state-of-the-art datasets. Its questions were not designed by predefined templates but were hand crafted by users, demonstrating high linguistic variance. Compared to previous datasets on knowledge bases it covers nearly 4,000 unique column headers, containing far more relations than closed domain datasets \cite{jagadish2014nalir,iyer2017learning} and datasets for querying knowledge bases \cite{cai2013large}.
Its questions cover a wide range of domains, requiring operations such as table lookup, aggregation, superlatives (argmax, argmin), arithmetic operations, joins and unions. The complexity of its questions can be shown in Tables \ref{table:wtq_examples} and \ref{table:test_query_comparison}.

The complete dataset contains 22,033 examples on 2,108 tables. As the \textit{test set}, 20\% of the tables and their associated questions were set aside, while the remaining tables and questions serve as the \textit{training set}. The separation between tables in the training and test sets forces the question answering system to handle new tables with previously unseen relations and entities.

% Training Feedback %
\subsection{Training on Feedback}\label{sec:trainfeedback}
The goal of the semantic parser is to translate natural language questions into equivalent formal queries. Thus, in order to ideally train the parser, we should train it on questions annotated with their respective queries. However, annotating NL questions with formal queries is a costly operation, hence recent works have trained semantic parsers on examples labeled solely with their answer \cite{clarke2010driving, liang2013learning, berant2013semantic, pasupat2015wtq}. This \textit{weak supervision} facilitates the training process at the cost of learning from incorrect queries. 
Figure \ref{fig:explanations_train} presents two candidate queries for the question \textit{"What was the last year the team was a part of the USL A-league?"}. Note that both queries output the correct answer to the question, which is \textit{2004}. However, the second query is clearly incorrect given its utterance is \textit{"minimum value in column Year in rows that have the highest value in column Open Cup"}.

The \textsc{WikiTableQuestions} dataset, on which the parser is trained, is comprised of question-answer pairs. Thus by retraining the parser on question-query pairs, that are provided as feedback, we can improve its overall correctness. We address this in our work by explaining queries to non-experts, enabling them to select the correct candidate query or mark \textit{None} when all are incorrect. 

These \textit{annotations} are then used to retrain the semantic parser. Given a question, its annotations are the queries marked as correct by users. We note that a question may have more than one correct annotation.

% Semantic Parsing %
\paragraph*{Semantic Parsing}\label{sec:semanticparsing}
Semantic Parsing is the task of mapping natural language questions to formal language queries (SQL, lambda DCS, etc.) that are executed against a target database. The semantic parser is a parameterized function, trained by updating its parameter vector such that questions from the training set are translated to formal queries yielding the correct answer.

We denote the table by $T$ and the NL question by $x$. The semantic parser aims to generate a query $z$ which executes to the correct answer of $x$ on $T$, denoted by $y$. In our running example from Figure \ref{figure:olympics}, the parser tries to generate queries which execute to the value \textit{2004}. 
We define $\mathcal{Z}_{x}$ as the set of candidate queries generated by parsing $x$. For each $z \in \mathcal{Z}_x$ we extract a feature vector $\phi(x,T,z)$ and define a log-linear distribution over candidates:
\begin{equation}
p_{\theta}(z|x,T) \propto \exp(\phi(x,T,z)^\top \theta)
\end{equation}
where $\theta$ is the parameter vector. We formally define the parser distribution of yielding the correct answer,
\begin{equation}
p_{\theta}(y|x,T) = \sum_{z\in \mathcal{Z}_{x}} r(z | T,y) \cdot p_{\theta}(z|x,T)
\end{equation}
where $r(z|y,T)$ is 1 when $z(T) = y$ and zero otherwise. 

The parser is trained using examples $\{(x_i, T_i, y_i)\}^N_{i=1}$, optimizing the parameter vector $\theta$ using AdaGrad \cite{duchi2011adaptive} in order to maximize the following objective \cite{pasupat2015wtq},
\begin{equation}
J(\theta) = \frac{1}{N}\sum^N_{i=1}\log p_\theta(y_i \mid x_i, T_i) + \lambda \|
\theta \|_1 
\label{eq:parser_objective}
\end{equation}
where $\lambda$ is a hyperparameter vector obtained from cross-validation.
To train a semantic parser that is unconstrained to any specific domain we deploy the parser in \cite{zhang2017macro}, trained end-to-end on the \textsc{WikiTableQuestions} dataset \cite{pasupat2015wtq}.

\paragraph*{Semantic Parsing on Annotations}
We modify the original parser so that \textit{annotated questions} are trained using \textit{question-query} pairs while all other questions are trained as before.
The set of annotated examples is denoted by $\mathcal{A}$. Given annotated example $x \in \mathcal{A}$, its set of valid queries is $\mathcal{Q}_{x}$. We define the distribution for an annotated example to yield the correct answer by,
\begin{equation}
p^{*}_\theta(y|x,T) = \sum_{z\in \mathcal{Z}_{x}}r^{*}(z|x,T) \cdot p_{\theta}(z|x,T)
\end{equation}
Where $r^{*}(z|x,T)$ is 1 when $z \in \mathcal{Q}_x$ and zero otherwise. Our new objective for retraining the semantic parser,
\begin{equation}
\begin{split}
J(\theta) = \frac{1}{|\mathcal{A}|}\sum_{x_i \in \mathcal{A}}\log p^{*}_\theta(y_i \mid x_i, T_i) \quad + \\ \frac{1}{N-|\mathcal{A}|}\sum_{x_i \notin \mathcal{A}}\log p_\theta(y_i \mid x_i, T_i)  \quad  + \quad  \lambda \|
\theta \|_1 
\end{split}
\end{equation}
the first sum denoting the set of annotated examples, while the second sum denotes all other examples. 

This enables the parser to update its parameters so that questions are translated into correct queries, rather than merely into queries that yield the correct answer.

\begin{figure}
	\vspace{-0.5em}
    \centering 
    \small
    \textbf{Question: What was the last year the team was a part of the USL A-League?}\newline
    \tiny
    
    \begin{minipage}{1\linewidth}
    	\scriptsize
    	\textbf{Utterance:} \textit{"maximum value in column Year in rows where value of column League is USL A-League"}\newline\newline
        \tiny
        \centering
        \begin{tabular}{|B>{\columncolor{highlight1}}c|B>{\columncolor{highlight1}}c|c|c|c|} \hline
		\textbf{\color{frame} MAX(\color{black}Year\color{frame})}  & \textbf{League}  &  \textbf{Attendance} &  \textbf{Open Cup} & ... \\ \hline
		\colorboxed{frame}{2002} & \colorboxed{frame}{USL A-League}  & 6,260 & Did not qualify & ...  \\ \hline
		\colorboxed{frame}{2003} & \colorboxed{frame}{USL A-League}  & 5,871 & Did not qualify & ...  \\ \hline
		\cellcolor{highlight2}\colorboxed{frame}{2004} & \colorboxed{frame}{USL A-League}  & 5,628 & 4th Round & ...  \\ \hline
		2005 & 	USL First Division  & 6,028 & 4th Round & ...  \\ \hline
		2006 & USL First Division  & 5,575 & 3rd Round & ... \\ \hline
        ... & ... & ... & ... & ... \\ \hline
		\end{tabular}
        \newline\newline
    \end{minipage}%

    \begin{minipage}{1\linewidth}
    	\scriptsize
    	\textbf{Utterance:} \textit{"minimum value in column Year in rows that have the highest value in column Open Cup"}\newline\newline
        \tiny
        \centering
        \begin{tabular}{|B>{\columncolor{highlight1}}c|c|c|B>{\columncolor{highlight1}}c|c|} \hline
		\textbf{\color{frame} MIN(\color{black}Year\color{frame})} & \textbf{League}  &  \textbf{Attendance} &  \textbf{Open Cup} & ... \\ \hline
		2002 & USL A-League  & 6,260 & Did not qualify & ...  \\ \hline
		2003 & USL A-League  & 5,871 & Did not qualify & ...  \\ \hline
		\cellcolor{highlight2}\colorboxed{frame}{2004} & USL A-League  & 5,628 & \colorboxed{frame}{4th Round} & ...  \\ \hline
		\colorboxed{frame}{2005} & 	USL First Division  & 6,028 & \colorboxed{frame}{4th Round} & ...  \\ \hline
		2006 & USL First Division  & 5,575 & 3rd Round & ... \\ \hline
        ... & ... & ... & ... & ... \\ \hline
		\end{tabular}
    \end{minipage}%    
    \caption{Correct \& incorrect query both returning the same answer}\label{fig:explanations_train}
    %\vspace{-5mm}
    \vspace{-1.5em}
\end{figure}
\newpage

% Deployment Feedback %
\subsection{Deployment}\label{sec:testfeedback}
At deployment, user interaction is used to ensure that the system returns formal-queries that are correct. 

We have constructed a web interface allowing users to pose NL questions on tables and by using our query explanations, to choose the correct query from the top-k generated candidates.
Normally, a semantic parser receives an NL question as input and displays to the user only the \textit{result} of its top ranked query. The user receives no explanation as to why was she returned this specific result or whether the parser had managed to correctly parse her question into formal language.
In contrast to the baseline parser, our system displays to users its top-k candidates, allowing them to modify the parser's top query.

\begin{example}
Figure \ref{fig:explanations_test} shows an example from the \textsc{WikitableQuestions} test set with the question \textit{"How many more ships were wrecked in lake Huron than in Erie"}. Note that the original table contains many more records than those displayed in the figure. 
Given the explanations of the parser's top candidates, our provenance-based highlights make it clear that the first query is correct as it compares the table occurrences of lakes Huron and Erie. The second result is incorrect, comparing lakes Huron and Superior, while the third query does not compare occurrences.
\end{example}

\begin{figure}
\vspace{-0.5em}
    \centering 
    \small
    \textbf{Question: How many more ships were wrecked in lake Huron than in Erie?}\newline\newline
    \tiny
    \begin{minipage}{1\linewidth}
    	\scriptsize
    	\textbf{Utterance:} \textit{"in column Lake, what is the difference between rows with value Lake Huron and rows with value Lake Erie"}
 \newline\newline
 		\tiny
        \centering
        \begin{tabular}{|c|c|B>{\columncolor{highlight1}}c|c|c|} \hline
		\textbf{Ship}  & \textbf{Vessel}  &  \textbf{Lake} &  \textbf{Lives lost} & ... \\ \hline
		Argus & Steamer & \cellcolor{diff1}\colorboxed{frame}{Lake Huron} & 25 lost & ... \\ \hline
		Hydrus & Steamer & \cellcolor{diff1}\colorboxed{frame}{Lake Huron} & 28 lost & ... \\ \hline
		Plymouth & Barge & Lake Michigan & 7 lost & ... \\ \hline
		Issac M. Scott & Steamer & \cellcolor{diff1}\colorboxed{frame}{Lake Huron} & 28 lost & ... \\ \hline
        Henry B. Smith & Steamer & Lake Superior & all hands & ...  \\ \hline
        Lightship No. 82 & Lightship & \cellcolor{diff2}\colorboxed{frame}{Lake Erie} & 6 lost & ... \\ \hline
        ... & ... & ... & ... & ... \\ \hline
		\end{tabular}
        \newline\newline
    \end{minipage}%
    
    \begin{minipage}{1\linewidth}
    	\scriptsize
    	\textbf{Utterance:} \textit{"in column Lake, what is the difference between rows with value Lake Huron and rows with value Lake Superior"}
 \newline\newline
 		\tiny
        \centering
        \begin{tabular}{|c|c|B>{\columncolor{highlight1}}c|c|c|} \hline
		\textbf{Ship}  & \textbf{Vessel}  &  \textbf{Lake} &  \textbf{Lives lost} & ... \\ \hline
		Argus & Steamer & \cellcolor{diff1}\colorboxed{frame}{Lake Huron} & 25 lost & ... \\ \hline
		Hydrus & Steamer & \cellcolor{diff1}\colorboxed{frame}{Lake Huron} & 28 lost & ...  \\ \hline
		Plymouth & Barge & Lake Michigan & 7 lost & ...  \\ \hline
		Issac M. Scott & Steamer & \cellcolor{diff1}\colorboxed{frame}{Lake Huron} & 28 lost & ...  \\ \hline
        Henry B. Smith & Steamer & \cellcolor{diff2}\colorboxed{frame}{Lake Superior} & all hands & ...  \\ \hline
        Lightship No. 82 & Lightship & Lake Erie & 6 lost & ...  \\ \hline
        ... & ... & ... & ... & ... \\ \hline
		\end{tabular}
        \newline\newline
    \end{minipage}%
    
    \begin{minipage}{1\linewidth}
    	\scriptsize
    	\textbf{Utterance:} \textit{"the number of rows where value of column Lake is Lake Huron that have the highest value in column Lives lost"}
 \newline\newline
 		\tiny
        \centering
        \begin{tabular}{|c|c|B>{\columncolor{highlight1}}c|B>{\columncolor{highlight1}}c|c|} \hline
		\textbf{Ship}  & \textbf{Vessel}  &  \textbf{\color{frame} COUNT(\color{black}Lake\color{frame})} &  \textbf{Lives lost} & ... \\ \hline
		Argus & Steamer & \colorboxed{frame}{Lake Huron} & 25 lost & ... \\ \hline
		Hydrus & Steamer & \cellcolor{highlight2}\colorboxed{frame}{Lake Huron} & \colorboxed{frame}{28 lost}  & ... \\ \hline
		Plymouth & Barge & Lake Michigan & 7 lost  & ... \\ \hline
		Issac M. Scott & Steamer & \cellcolor{highlight2}\colorboxed{frame}{Lake Huron} & \colorboxed{frame}{28 lost}  & ... \\ \hline
        Henry B. Smith & Steamer & Lake Superior & all hands  & ... \\ \hline
        Lightship No. 82 & Lightship & Lake Erie & 6 lost  & ... \\ \hline
        ... & ... & ... & ... & ... \\ \hline
		\end{tabular}
    \end{minipage}%
    
    \caption{Identifying the correct query through provenance-based highlights}\label{fig:explanations_test}
\vspace{-2em}
\end{figure}

% Experiments %
\section{Experiments}\label{sec:experiments}
Following the presentation of concrete applications for our methods we have designed an experimental study to measure the effect of our query explanation mechanism. We conducted experiments to evaluate both the quality of our explanations, as well as their contribution to the baseline parser. This section is comprised of two main parts:

\begin{itemize}
  \item \textbf{Interactive Parsing:} We have deployed our NL interface online, explaining to users the top candidate queries generated for over 400 distinct questions. Our user study measured the impact of query explanations on choosing correct queries while also comparing the average work-time of users with and without our provenance-based highlights.
  \item \textbf{Training on Feedback:} We stored user feedback as question-query pairs, and used it to retrain the system's semantic parser. Experiments show an increase in parser correctness when trained on user feedback.
\end{itemize}

The experimental results show our query explanations to be effective, allowing non-experts to easily understand generated queries and to disqualify incorrect ones. Training on user feedback further improves the system correctness, allowing it to learn from user experience.

% Evaluation Metrics %
\subsection{Evaluation Metrics}
We begin by defining the system \textit{correctness}, used as our main evaluation metric.
Recall that the semantic parser is given an NL question $x$ and table $T$ and generates a set $\mathcal{Z}_{x}$ of candidate queries. Each query $z \in \mathcal{Z}_{x}$ is then executed against the table, yielding result $z(T)$. We define the parser \textit{correctness} as the percentage of questions where the \textit{top-ranked} query is a \textit{correct translation} of $x$ from NL to lambda DCS. In addition to correctness, we also measured the \textit{mean reciprocal rank} (MRR), used for evaluating the average correctness of all candidate queries generated, rather than only that of the top-1.

\begin{example}
To illustrate the difference between correct answers and correct queries let us consider the example in Figure \ref{fig:explanations_train}. The parser generates the following candidate queries (we present only their utterances):
\begin{itemize}
  \item \textit{maximum value in column Year in rows where value of column League is USL A-League}.
  \item  \textit{minimum value in column Year in rows that have the highest value in column Open Cup}.
\end{itemize}
Both return the correct answer \textit{2004}, however only the first query conveys the \textit{correct} translation of the NL question.
\end{example}

% Interactive Parsing at Deployment %
\subsection{Interactive Parsing at Deployment}\label{sec:experiments_deployment}
We use query explanations to improve the real-time performance of the semantic parser. Given any NL question on a (never before seen) table, the parser will generate a set of candidate queries. Using our explanations, the user will interactively select the correct query (when generated) from the parser's top-k results. We compare the correctness scores of our interactive method with that of the baseline parser.

\paragraph*{User Study} Our user study was conducted using anonymous workers recruited through the the Amazon Mechanical Turk (AMT) crowdsourcing platform. Focusing on non-experts, our only requirements were that participants be over 18 years old and reside in a native English speaking country. Our study included 35 distinct workers, a significant number of participants compared to previous works on NL interfaces \cite{deutch2017nlprov,jagadish2014nalir,koutrika2010explaining}.
Rather than relying on a small set of NL test questions \cite{deutch2017nlprov,jagadish2014nalir} we presented each worker with 20 distinct questions that were randomly selected from the \textsc{WikiTableQuestions} benchmark dataset (Section \ref{sec:wtq}). A total of 405 distinct questions were presented (as described in Table \ref{table:success}). For each question, workers were shown explanations (utterances, highlights) of the top-7 candidate queries generated. Candidates were randomly ordered, rather than ranked by the parser scores, so that users will not be biased towards the parser's top query. Given a question, participants were asked to mark the correct candidate query, or \textit{None} if no correct query was generated.

Displaying the top-k results allowed workers to improve the baseline parser in cases where the correct query was generated, but not ranked at the top. After examining different values of $k$, we chose to display top-k queries with $k=7$. We made sure to validate that our choice of $k=7$ was sufficiently large, so that it included the correct query (when generated). We randomly selected 100 examples where no correct query was generated in the top-7 and examined whether one was generated within the top-14 queries.
Results had shown that for $k=14$ only 5\% of the examples contained a correct query, a minor improvement at the cost of doubling user effort. Thus a choice of $k=7$ appears to be reasonable.

\paragraph*{User Success} To verify that our query explanations were understandable to non-experts we measured each worker's success. Results in Table \ref{table:success} show that in 78.4\% of the cases, workers had succeeded in identifying the correct query or identifying that no candidate query was correct. The average success rate for all 35 workers being 15.7/20 questions. When comparing our explanation approach (utterances + highlights) to a baseline of \textit{no explanations}, non-expert users failed to identify correct queries when shown only lambda DCS queries. This demonstrates that utterances and provenance-based highlights serve as effective explanations of formal queries to the layperson. We now show that using them jointly is superior to using only utterances.

\begin{table}[t]
\centering
\small
\caption{User Study - Success Rates}
\begin{tabular}{|c|c|c|} \hline
\textbf{distinct questions}  & \textbf{explanations}  &  \textbf{avg. success}\\ \hline
405 & 2,835 & 78.4\% \\ \hline
\end{tabular}
\scriptsize{\newline\newline The number of distinct questions and candidate query explanations presented to users. Users successfully identified 78.4\% of the explanations as being correct or incorrect.}
\label{table:success}
\end{table}

\paragraph*{Work-time Results} When introducing our two explanation methods, we noted their complementary nature. NL utterances serve as highly detailed phrases describing the query, while highlighting provenance cells allows to quickly single out the correct queries. We put this claim to the test by measuring the impact our novel provenance-based highlights had on the average work-time of users. 
We measured the work-time of 20 distinct AMT workers, divided into two separate groups, each containing half of the participants. Workers from both groups were presented with 20 questions from \textsc{WikiTableQuestions}. The first group of workers were presented both with highlights and utterances as their query explanations, while the second group had to rely solely on NL utterances.
Though both groups achieved identical correctness results, the group employing table highlights performed significantly faster. Results in Table \ref{table:worktime} show our provenance-based explanations cut the average and median work-time by 34\% and 20\% respectively. Since user work-time is valuable, the introduction of visual explanations such as table highlights may lead to significant savings in worker costs.

\begin{table}[t]
\centering
\small
\caption{User Work-Time (minutes) on 20 questions}
\begin{tabular}{|c|c|c|c|c|} \hline
\textbf{method} & \textbf{avg} & \textbf{median}& \textbf{min}& \textbf{max}\\ \hline
Utterances + Highlights & 16.2m & 16.6m & 6.45m & 22.5m\\ \hline
Utterances & 24.7m & 20.7m & 17.5m & 35.4m\\ \hline
\end{tabular}
\label{table:worktime}
\end{table}

\paragraph*{Correctness Results}
We have examined the effect to which our query explanations can help users improve the correctness of a baseline NL interface.   
Our user study compares the correctness of three scenarios:
\begin{itemize}
  \item \textit{Parser correctness} - our baseline is the percentage of examples where the top query returned by the semantic parser was correct.
  \item \textit{User correctness} - the percentage of examples where the user selected a correct query from the top-7 generated by the parser.
  \item \textit{Hybrid correctness} - correctness of queries returned by a combination of the previous two scenarios. The system returns the query marked by the user as correct; if the user marks all queries as incorrect it will return the parser's top candidate.
\end{itemize}

Results in Table \ref{table:test} show the correctness rates of these scenarios. User correctness score is superior to that of the baseline parser by 7.5\% (from 37.1\% to 44.6\%), while the hybrid approach outscores both with a correctness of 48.7\% improving the baseline by 11.6\%. For the user and hybrid correctness we used a $\chi^2$ test to measure significance. Random queries and tables included in the experiment are presented in Table \ref{table:test_query_comparison}. We also include a comparison of the top ranked query of the baseline parser compared to that of the user.

We define the \textit{correctness bound} as the percentage of examples where the top-k candidate queries actually contain a correct result. This bound serves as the optimal correctness score that workers can achieve. The 56\% correctness-bound of the baseline parser stems from the sheer complexity of the \textsc{WikiTableQuestions} benchmark. Given the training and test tables are disjoint, the parser is tested on relations and entities unobserved during its training. This task of generalizing to unseen domains is an established challenge in semantic parsing \cite{pasupat2015wtq,zhong2017seq2sql}.
Using the correctness-bound as an upper bound on our results shows the hybrid approach achieves 87\% of its full potential. Though there is some room for improvement, it seems reasonable given that our non-expert workers possess no prior experience of their given task.

\paragraph*{Execution Times} We describe the execution times for generating our query explanations in Table \ref{table:exectime}. We trained the semantic parser using the SMEPRE toolkit \cite{berant2013semantic} on a machine with Xeon 2.20GHz CPU and 256GB RAM running Linux Ubuntu 14.04 LTS. We report the average generation times of candidate queries, utterances and highlights over the entire \textsc{WikiTableQuestions} test set, numbering 4,344 questions.

\begin{table}[t]
\centering
\small
\caption{User Study - Correctness Results}
\begin{tabular}{|c|c|c|} \hline
  & \textbf{correct examples}  &  \textbf{correctness}\\ \hline
Parser & 260/700 & 37.1\% \\ \hline
Users & 312/700$\dagger$ & 44.6\% \\ \hline
Hybrid & \textbf{341}/700$\dagger$ & \textbf{48.7\%} \\ \hline\hline
Bound & 392/700 & 56\% \\ \hline
\end{tabular}
\scriptsize{\newline\newline The $\dagger$ sign indicates statistical significance compared to the Parser baseline at the 0.01 level using the $\chi^2$ test (with 1 degree of freedom).}
\label{table:test}
\end{table}

\begin{table}[t]
\centering
\small
\caption{Avg. Execution Time (seconds) on \textsc{WikiTableQuestions}}
\begin{tabular}{|c|c|c|c|} \hline
\textbf{Questions} & \textbf{Cand. Gen.} & \textbf{Utter. Gen.}& \textbf{Highlights Gen.} \\ \hline
4,344 & 1.22 & 0.22 & 1.36 \\ \hline
\end{tabular}
\label{table:exectime}
\end{table}

\begin{table*}
\scriptsize
\centering
\caption{User Study - Questions and Answers}
\begin{tabular}{p{4.2cm} p{2.8cm} p{4.4cm} p{4.4cm} } \hline
  \textbf{Question}  &  \textbf{Table Attributes} & \textbf{User explanation choice}  &  \textbf{Parser baseline}  \\ \hline
What is the difference in number of defensive player of the year awards received by Gabriel Gervais and Mauricio Vincello? & Year, MVP, Defensive Player of the Year, Unsung Hero, Newcomer of the Year & \textit{in column Defensive Player of the Year, what is the difference between rows with value Mauricio Vincello and rows with value Gabriel Gervais} & \textit{in column Defensive Player of the Year, what is the difference between rows with value Mauricio Vincello and rows with value Mauricio Vincello} \\ \hline
The next European team Haiti played after Spain on June 8, 2013 was which team? & Date, Location, Opponent, Result, Competition & \textit{values in column Opponent in rows right below rows where value of column Date is June 8 2013} & \textit{values in column Opponent in rows right below rows where value of column Opponent is June 8 2013 or Spain} \\ \hline
Who superseded lord high steward? & Position, Officer, Current Officers, Superseded By, Royal Household & \textit{values in column Superseded By in rows where value of column Officer is lord high steward} & \textit{values in column Royal Household in rows where value of column Officer is lord high steward} \\ \hline
Who has won the least amount of medals? & Rank, Nation, Gold, Silver, Bronze, Total& \textit{values in column Nation in rows that have the lowest value in column Total} & \textit{between values in column Nation in rows, who has the lowest value of column Bronze out of the values in Nation} \\ \hline
Who is older, Tatiana Abramenko or Myriam Asfry? & Represent, Candidate, in Russian, Age, Height, Hometown & \textit{between Myriam Asfry or Tatiana Abramenko, who has the highest value of column Age out of the values in Candidate} & \textit{between Myriam Asfry or Tatiana Abramenko, who has the lowest value of column Age out of the values in Candidate} \\ \hline
The title of the last show was? & No., Episode, Air date, Rating, Share, Viewers & \textit{between values in column Episode in rows, who has the highest value of column Air date out of the values in Episode} & \textit{between values in column No. in rows, who has the highest value of column Air date out of the values in No.} \\ \hline
\end{tabular}
\label{table:test_query_comparison}
\end{table*}

% Training on User Feedback %
\subsection{Training on User Feedback}
We measure our system's ability to learn from user feedback in the form of question-query pairs. Given a question, the user is shown explanations of the parser's top-7 queries, using them to \textit{annotate} the question, i.e. assign to it correct formal queries (e.g., the first query in Figure \ref{fig:explanations_train}). 
Annotations were collected by displaying users with questions from the \textsc{WikiTableQuestions} training set along with query explanations of the parser results.
To enhance the annotation quality, each question was presented to three distinct users, taking only the annotations marked by at least two of them as correct.
Data collection was done using AMT and in total, 2,068 annotated questions were collected. 
Following a standard methodology, we split the annotated data into train and development sets. Out of our 2,068 annotated examples, 418 were selected as the development set, and 1,650 as the training set.
The annotated development examples were used to evaluate the effect of our annotations on the parser correctness.

We experiment on two scenarios: 
(1) training the parser solely on 1,650 annotated examples; 
(2) integrating our training examples into the entire \textsc{WikiTableQuestions} training set of 11K examples.
For each scenario we trained two parsers, one trained using annotations and the other without any use of annotations.
To gain more robust results we ran our experiments on three different train/dev splits of our data, averaging the results.
Table \ref{table:correctness} displays the results of our experiments.
When training solely on the annotated examples, parser correctness on development examples increased by 8\% (41.8\% to 49.8\%). The spike in correctness shows that feedback acquired using our explanations is high-quality input for the semantic parser, hence the parser achieves better correctness when trained on it compared to training on the original \textsc{WikiTableQuestions} benchmark.

When training on all 11K train examples using our 1,650 annotations we also saw an increase (of 2.1\%), albeit being more modest due to the percentage of annotated examples.
We witnessed an increase in both correctness and MRR (mean reciprocal rank) that grows in the number of annotated train examples. 
This further asserts the significance of annotated training data \cite{poon2013grounded, yih2016value} and shows that our system can learn from quality feedback collected by non-experts.

\begin{table}
\centering
\small
\caption{Effect of user feedback on correctness}
\begin{tabular}{|c|c|c|c|} \hline
\textbf{train ex.} & \textbf{annotations} & \textbf{correctness} & \textbf{MRR} \\ \hline
1650 & 1650 & \textbf{49.8\%} & \textbf{0.586} \\ \hline
1650 & 0 & 41.8\% & 0.499 \\ \hline\hline
11000 & 1650 & \textbf{51.6\%} & \textbf{0.60} \\ \hline
11000 & 0 & 49.5\% & 0.570 \\ \hline
\end{tabular}
\label{table:correctness}
\end{table}

% Related Work %
\section{Related Work}\label{sec:related}
\paragraph*{NL interfaces}
Building a natural language interface (NLIDB) for querying databases has been extensively studied in the literature \cite{kuepper1993nauda, androutsopoulos1995natural, popescu2003towards,giordani2012translating, jagadish2014nalir, iyer2017learning}. Notably, \textsc{NaLIR} \cite{jagadish2014nalir} is an NLIDB where users are presented explanations of candidate queries in the form of an intermediate representation, termed the \textit{query tree}. Since we are geared towards non-experts, instead of a query tree representation, we explain parsed queries by highlighting their cell-based provenance while also providing detailed NL utterances.

Explaining formal queries in NL has been studied both on relational database schemas \cite{koutrika2010explaining,analyza2017} and KB systems \cite{wang2015overnight}. Our domain independent utterances are comparable to the generic template of \cite{koutrika2010explaining}, however when challenged with complex queries, both methods are forced to return utterances which are quite long. The paper authors solve this by having users manually devise schema specific templates. We are able to leverage our adherence to single tables by providing provenance-based highlights to visually explain complex queries to non-experts. The evaluation of our query explanations was focused on the comprehension of \textit{non-experts}, in contrast to the expert driven evaluation presented in \cite{koutrika2010explaining}. Our user study empirically showed a joint approach of NL utterances and table highlights facilitates user understanding compared to explaining only through utterances \cite{analyza2017}. 

NL utterances have also been used in the context of data exploration systems \cite{clustine2016,analyza2017}. The work in \cite{clustine2016} compresses user query \textit{results} and describes the resulting clusters in NL. We focus on an orthogonal task of explaining formal queries, in order improve a state-of-the-art semantic parser.

Machine learning systems for querying knowledge bases have long been the standard in NLP research. From systems for question answering over \textsc{Freebase} \cite{berant2013semantic, yih2016value}, to NL interfaces mapping directly to SQL \cite{iyer2017learning,zhong2017seq2sql}. The work in \cite{zhang2017macro} presents a state-of-the-art semantic parser over the NLP community benchmark of \textsc{WikiTableQuestions} \cite{pasupat2015wtq}. We further improve the correctness of the parser by integrating our query explanations into an NL interface over this dataset.
The usage of procured user feedback to train semantic parsers was also pursued in recent works \cite{yih2016value, iyer2017learning}. However, both works rely on additional labeling by expert users (familiar in SQL, SPARQL), while our parser is trained solely from the feedback of its non-expert users. Our approach of utilizing user feedback to enhance the interface both at deployment and also in retraining the parser is a joint implementation of both database and NLP common practices \cite{jagadish2014nalir, iyer2017learning,yih2016value}.

\paragraph*{Provenance}
Provenance models have long been studied in the context of relational queries \cite{buneman2001and, cheney2009provenance, davidson2008provenance, glavic2013using, green2011containment, herschel2016provenance}. The complexity of provenance expressions resulted in multiple approaches to represent provenance that is user-understandable. These include provenance in a graph form \cite{ailamaki1998scientific, davidson2008provenance, foster2002chimera, oinn2004taverna, simmhan2010karma2} and methods that present different ways of provenance visualization \cite{herschel2016provenance}.  The work in \cite{deutch2017nlprov} presents \textsc{NLProv}, an interface built on top of \textsc{NaLIR} complete with a provenance model for NL queries. Their solution however, is limited to handling Conjunctive Queries. The cell-based provenance we present supports further NL constructs such as union, aggregation, arithmetic difference and more. Our model is able to explain NL questions that show high diversity in structure and linguistic compositionality, tested on a benchmark NLP dataset for question answering over tables \cite{pasupat2015wtq}.

% Conclusion %
\section{Conclusion and Future Work}\label{sec:conclusion_future_work}
We have studied in this paper the problem of explaining complex NL queries to non expert users. We introduced visual query explanations in the form of table highlights, based on a novel cell-based provenance model tested on web tables from hundreds of distinct domains. Table highlights provide immediate visual feedback for identifying correct candidate queries. We combine table highlights with utterance based query explanations, significantly improving their effectiveness.
Using our query explanations we enhanced an NL interface for querying tables by providing it with feedback at both deployment and training time. Feedback is procured through query explanations, allowing users with no technical background to query tables with confidence, while simultaneously providing feedback to enhance the interface itself. We implement a \textit{human in the loop} paradigm, where our users both exploit the underlying Machine Learning algorithm while providing it with further data to train on.

We have put our methods to the test, having conducted an extensive user study to determine the clarity of our explanations. Experimenting with explanations for hundreds of formal queries, users proved to be successful in interactively choosing correct queries, easily topping the baseline parser correctness. The addition of provenance-based highlights helps boost the efficacy of user feedback, cutting average work-time by a third compared to the utterances baseline. 

\paragraph*{Future Work}
In our implementation, the retraining of the parser on user annotations is performed offline. In future work we aim to enable the parser to learn from users at run-time, via online learning techniques. Instead of asking the user to choose a query from the top-k results, or mark all of them as incorrect, an online parser may \textit{query the user} until the correct query is generated. Such a system should be expected to learn interactively whether to return its top-ranked query, or seek further clarifications from the user.

Additional work might explore the impact of user understandable query explanations on semantic parsing tasks lacking \textit{strong supervision}. As we described in Section \ref{sec:trainfeedback} annotating questions with formal queries is a costly operation, hence recent works have trained parsers on questions labeled solely with their answer \cite{clarke2010driving, liang2013learning, berant2013semantic, pasupat2015wtq}.
By explaining the candidate queries of a baseline semantic parser non experts workers were able to annotate thousands of examples from the \textsc{WikiTableQuestions} dataset without any knowledge of the formal language in use.
Explaining semantic parser candidate queries can be employed on other datasets, allowing non-experts to easily annotate significant amounts of the data.
The positive impact of query annotations has been established in \cite{yih2016value}, hence by annotating \textit{weakly supervised} datasets we will produce quality training data used to improve the accuracy of state-of-the-art parsers on a myriad of tasks.

Another direction worth exploring is attempting to significantly improve the \textsc{WikiTableQuestions} benchmark through use of our collected annotations and the introduction of new neural sequence-to-sequence models replacing the parser of \cite{zhang2017macro} that we used. Neural models have shown great promise in semantic parsing tasks when trained on question-query pairs \cite{iyer2017learning,zhong2017seq2sql}. We hope to improve benchmark results by using our approach to annotate most of \textsc{WikiTableQuestions} then a training state-of-the-art neural sequence-to-sequence on the collected data.

\begin{table*}
\scriptsize 
\centering
\caption{Lambda DCS Operators, SQL Translation and Provenance}
\begin{tabular}
{|p{1.5cm}|p{2.5cm}|p{3cm}|p{3.7cm}|p{5.5cm}|}\hline
\textbf{Operator} & \textbf{Query (lambda DCS)} & \textbf{Example} & \textbf{Semantics (SQL)} & \textbf{Provenance}\\ \hline
Column Records & \texttt{C.v} & \texttt{City.Athens} & \texttt{SELECT * FROM T\newline WHERE C = 'v';} & $P_O(Q) = \{c \mid c \in C \; \land \; T[c] = v\}$ \newline$P_E(Q) = P_O(Q)$\newline
$P_C(Q) = \{c \mid c \in C\}$\\ \hline
Column Values & \texttt{\textbf{R}[C].records} & \texttt{\textbf{R}[Year].City.Athens} & \texttt{SELECT C FROM (records);} & $P_O(Q) = \{c \mid c \in C \; \land \; record(c).Index \in records.Indices\} $ \newline $P_E(Q) = P_O(Q) \cup P_E(records) $ \newline $P_C(Q) = \{c \mid c \in C\} \cup P_C(records)$\\ \hline
Values in Preceding Records & \texttt{R[C].Prev.records} & \texttt{R[Year].Prev.City.\newline Athens} & \texttt{SELECT C FROM T \newline WHERE Index IN (\newline\hspace*{0.7cm} SELECT Index-1 \newline\hspace*{0.7cm} FROM (records));} & $P_O(Q) = \{c \mid c \in C \; \land \; record(c).Index+1 \in records.Indices\} $ \newline $P_E(Q) = P_O(Q) \cup P_E(records) $ \newline $P_C(Q) = \{c \mid c \in C\} \cup P_C(records)$\\ \hline
Values in Following Records & \texttt{R[C].R[Prev].\newline records} & \texttt{R[Year].R[Prev].City.\newline Athens} & \texttt{SELECT C FROM T \newline WHERE Index IN (\newline\hspace*{0.7cm} SELECT Index+1 \newline\hspace*{0.7cm} FROM (records));} & 
$P_O(Q) = \{c \mid c \in C \; \land \; record(c).Index-1 \in records.Indices\} $ \newline $P_E(Q) = P_O(Q) \cup P_E(records) $ \newline $P_C(Q) = \{c \mid c \in C\} \cup P_C(records)$\\ \hline
Aggregation on Values & \texttt{aggr(vals)} \newline \textcolor{gray}{aggr $\in$ \{count, max, min, sum, avg\}} & \texttt{sum(R[Year].City.\newline Athens)} & \texttt{SELECT AGGR(C) FROM (vals);} & $P_O(Q) = P_O(vals) \cup \{AGGR\} $ \newline $P_E(Q) = P_O(Q)$ \newline $P_C(Q) = \{c \mid c \in C \}$\\ \hline
Difference of Values & \texttt{sub(\textbf{R}[C1].C2.v, \textbf{R}[C1].C2.u)} & \texttt{sub(\newline \textbf{R}[Year].City.London, \textbf{R}[Year].City.Beijing)} & \texttt{(\newline SELECT C1 FROM T\newline WHERE Index IN(\newline \hspace*{0.7cm} SELECT Index FROM T\newline \hspace*{0.7cm} WHERE C2 = 'v'))\newline- (\newline SELECT C1 FROM T\newline WHERE Index IN(\newline \hspace*{0.7cm} SELECT Index FROM T\newline \hspace*{0.7cm} WHERE C2 = 'u'));} & $P_O(Q) = P_O(R[C_1].C_2.v) \cup P_O(R[C_1].C_2.u)$ \newline $P_E(Q) = P_E(R[C_1].C_2.v) \cup P_E(R[C_1].C_2.u)$ \newline $P_C(Q) = \{c \mid c \in C_1 \; \lor \; c \in C_2\}$\\ \hline
Difference of Value Occurrences & \texttt{sub(count(C.v), count(C.u))} & \texttt{sub(\newline count(City.Athens), count(City.London))} & \texttt{(\newline SELECT COUNT(Index) FROM T \newline WHERE C = 'v') \newline- (\newline SELECT COUNT(Index) FROM T \newline WHERE C = 'u');} &  $P_O(Q) = P_O(C.v) \cup P_O(C.u)$ \newline $P_E(Q) = P_E(C.v) \cup P_E(C.u)$ \newline $P_C(Q) = \{c \mid c \in C\}$\\ \hline
Union of Values & \texttt{vals\textsubscript{1} $\sqcup$ vals\textsubscript{2}} &\texttt{Country.China $\sqcup$ Country.Greece} & \texttt{SELECT C FROM (vals\textsubscript{1})\newline UNION \newline SELECT C FROM (vals\textsubscript{2});} &  $P_O(Q) = P_O(vals_1) \cup P_O(vals_2) $ \newline $P_E(Q) =  P_E(vals_1) \cup P_E(vals_2)$ \newline $P_C(Q) = \{c \mid c \in C \}$\\ \hline
Intersection of Records & \texttt{records\textsubscript{1} $\sqcap$ records\textsubscript{2}} & \texttt{City.London $\sqcap$ Country.UK} & \texttt{SELECT * FROM T \newline WHERE Index IN (\newline \hspace*{0.7cm} (SELECT Index FROM records\textsubscript{1}) \newline AND Index IN \newline \hspace*{0.7cm} (SELECT Index FROM records\textsubscript{2}));} & $P_O(Q) = P_O(records_1) \cap P_O(records_2) $ \newline $P_E(Q) = P_E(records_1) \cup P_E(records_2)$ \newline $P_C(Q) = P_C(records_1) \cup P_C(records_2)$\\ \hline
Records with Highest Value & \texttt{argmax(Record, $\lambda$x[C.x])} & \texttt{argmax(Record, $\lambda$x[Year.x])} & \texttt{SELECT * FROM T \newline WHERE C = (\newline \hspace*{0.7cm} SELECT MAX(C) FROM T );} & $P_O(Q) = \{c \mid c \in C \; \land \; \forall c' \in C. \; T[c'] \leq T[c]\} $ \newline $P_E(Q) =  P_O(Q)$ \newline $P_C(Q) = \{c \mid c \in C \}$\\ \hline
Value in Record with Highest Index & \texttt{R[C].argmax(\newline records, Index)} & \texttt{R[Year].argmax(City.\newline Athens, Index)} & \texttt{SELECT C FROM T \newline WHERE Index = (\newline\hspace*{0.7cm} SELECT MAX(Index) \newline\hspace*{0.7cm} FROM (records));} & $P_O(Q) = \{c \mid \forall r \in records \; r.Index \leq record(c).Index \} $ \newline $P_E(Q) =  P_O(Q) \; \cup \; P_E(records)$ \newline $P_C(Q) = \{c \mid c \in C \} \; \cup \; P_C(records)$\\ \hline
Value with Most Appearances & \texttt{argmax(vals, \textbf{R}[$\lambda$x.count(C.x)])} & \texttt{argmax(Athens $\sqcup$ London, \textbf{R}[$\lambda$x.count(City.x)])} & \texttt{SELECT C FROM T\newline WHERE Index = (\newline \hspace*{0.7cm} SELECT COUNT(Index) \newline\hspace*{0.7cm} FROM T \newline\hspace*{0.7cm} WHERE C IN (vals) \newline\hspace*{0.7cm} GROUP BY C \newline\hspace*{0.7cm} ORDER BY COUNT(Index) DESC\newline\hspace*{0.7cm} LIMIT 1);} & $P_O(Q) = \{c \mid c \in C \; \land \; \forall c' \in C \; COUNT(C.T[c']) \leq COUNT(C.T[c])\} $ \newline $P_E(Q) =  P_O(Q) \; \cup \; P_E(vals)$ \newline $P_C(Q) = \{c \mid c \in C \} \; \cup \; P_C(vals)$\\ \hline
Comparing Values & \texttt{argmax(vals, \textbf{R}[$\lambda$x.\textbf{R}[C1].C2.x])} & \texttt{argmax(London $\sqcup$ Beijing, \textbf{R}[$\lambda$x.\textbf{R}[Year].City.x])} & \texttt{SELECT DISTINCT C2 FROM T \newline WHERE C1 = (\newline \hspace*{0.7cm} SELECT MAX(C1) FROM T \newline \hspace*{0.7cm} WHERE C2 IN (vals)); } & $P_O(Q) = \{c_2 \mid c_2 \in C_2 \; \land \; T[c_2] \in vals \; \land \; \exists c_1 \in C_1. \; record(c_1) \equiv records(c_2) \; \land \; \forall c'_2 \in C_2. \; \exists c'_1 \in C_1. \; (record(c'_1) \equiv record(c'_2) \rightarrow T[c'_1] \leq T[c_1]) \} $ \newline $P_E(Q) =  P_E(vals) \; \cup \; \{c_1 \mid c_1 \in C_1 \; \land \; \exists c_2 \in C_2. \; record(c_1) \equiv record(c2) \; \land \; T[c_2] \in vals \}$ \newline $P_C(Q) = \{c \mid c \in C_1 \; \lor \; c \in C_2\}$\\ \hline

\end{tabular}
\label{table:lambdaDCS}
\end{table*}

\begin{table*}[!htp]
    \centering \scriptsize
    \begin{minipage}{.5\linewidth}
        \centering
        \begin{tabular}{|B>{\columncolor{highlight1}}c | c | c | c | c | c |}
            \hline \textbf{Name} & \textbf{Type} & \textbf{Owner} & ... \\
            \hline Sally & Yacht & Lyman & ... \\
            \hline Caprice & Yacht & Robinson & ... \\
            \hline Eleanor & Yacht & Clapp & ... \\
            \hline USS Lawrence & Yacht & U.S. Navy & ...\\
            \hline USS Macdonough & Yacht & U.S. Navy & ...\\
            \hline \cellcolor{highlight2}\colorboxed{frame}{Jule} & Yacht & J. Arthur & ... \\
            \hline lightship LV-72 & Lightvessel & U.S Lighthouse Board & ...\\
            \hline ... & ... & ... & ...\\
            \hline
        \end{tabular}
        \caption{Simple Join}\label{tbl:highlight_join}
        \small
        \textit{rows where value of column Name is Jule.}
    \end{minipage}%
    \begin{minipage}{.5\linewidth}
        \centering
        \begin{tabular}{|c | c |B>{\columncolor{highlight1}} c | c | c | c |}
            \hline \textbf{Name} & \textbf{Position} & \textbf{Games} & \textbf{Club} & ...\\
            \hline Erich Burgener & GK & 3 & Servette & ...\\
            \hline Roger Berbig & GK & 3 & Grasshoppers & ...\\
            \hline Charly In-Albon & DF & 4 & Grasshopers & ... \\
            \hline Beat Rietmann & DF & 2 & FC St. Gallen & ...\\
            \hline Andy Egli & DF & \cellcolor{highlight2}\colorboxed{frame}{6} & Grasshoppers & ...\\
            \hline Marcel Koller & DF & 2 & Grasshoppers & ...\\
            \hline Rene Botteron & MF & 1 & FC Nuremburg & ...\\
            \hline Heinz Hermann & MF & \cellcolor{highlight2}\colorboxed{frame}{6} & Grasshoppers & ...\\
            \hline Roger Wehrli & MF & \cellcolor{highlight2}\colorboxed{frame}{6} & Grasshoppers & ...\\
            \hline Lucien Favre & MF & \cellcolor{highlight2}\colorboxed{frame}{5} & Toulouse Servette & ...\\
            \hline ... & ... & ... & ...\\
            \hline
    %\hspace*{-0.4cm}
        \end{tabular}
        \caption{Comparison}\label{tbl:highlight_comparison}
        \small
        \textit{rows where values of column Games are more than 4.\newline}
    \end{minipage}

    \begin{minipage}{.33\linewidth}
        \centering
        \begin{tabular}{|B>{\columncolor{highlight1}} c | c |B>{\columncolor{highlight1}} c | c | c | c |}
            \hline \textbf{Year} & \textbf{Country} & \textbf{City} & ...\\ \hline
			\cellcolor{highlight2}\colorboxed{frame}{1896} & Greece & \colorboxed{frame}{Athens} & ...\\ \hline
			1900 & France & Paris & ...\\ \hline
			... & ... & ... & ...\\ \hline
			\cellcolor{highlight2}\colorboxed{frame}{2004} & Greece & \colorboxed{frame}{Athens} & ...\\ \hline
			2008 & China & Beijing & ...\\ \hline
			2012 & UK & London & ...\\ \hline
            2016 & Brazil & Rio de Janeiro & ...\\ \hline
        \end{tabular}
        \caption{Reverse Join}\label{tbl:highlight_reverse_join}
        \small
        \textit{values of column Year in rows where value of column City is Athens.}\newline
    \end{minipage}%
    \begin{minipage}{.33\linewidth}
        \centering
        \begin{tabular}{|c | c |B>{\columncolor{highlight1}} c | c | c | c |}
            \hline \textbf{Year} & \textbf{Country} & \textbf{City} & ...\\ \hline
			1896 & Greece & Athens & ...\\ \hline
			1900 & France & Paris & ...\\ \hline
			... & ... & ... & ...\\ \hline
			2004 & Greece & Athens & ...\\ \hline
			2008 & China & \cellcolor{highlight2}\colorboxed{frame}{Beijing} & ...\\ \hline
			2012 & UK & \colorboxed{frame}{London} & ...\\ \hline
            2016 & Brazil & Rio de Janeiro & ...\\ \hline
    %\hspace*{-0.4cm}
        \end{tabular}
        \caption{Previous}\label{tbl:highlight_next}
        \small
        \textit{values of column City right above rows where values of column City is London.\newline}
    \end{minipage}
    \begin{minipage}{.33\linewidth}
        \centering
        \begin{tabular}{|c | c |B>{\columncolor{highlight1}} c | c | c | c |}
            \hline \textbf{Year} & \textbf{Country} & \textbf{City} & ...\\ \hline
			1896 & Greece & \colorboxed{frame}{Athens} & ...\\ \hline
			1900 & France & \cellcolor{highlight2}\colorboxed{frame}{Paris} & ...\\ \hline
			... & ... & ... & ...\\ \hline
			2004 & Greece & \colorboxed{frame}{Athens} & ...\\ \hline
			2008 & China & \cellcolor{highlight2}\colorboxed{frame}{Beijing} & ...\\ \hline
			2012 & UK & London & ...\\ \hline
            2016 & Brazil & Rio de Janeiro & ...\\ \hline
    %\hspace*{-0.4cm}
        \end{tabular}
        \caption{Next}\label{tbl:highlight_prev}
        \small
        \textit{values of column City right below rows where values of column City is Athens.\newline}
    \end{minipage}

    \begin{minipage}{.33\linewidth}
        \centering
        \begin{tabular}{|B>{\columncolor{highlight1}} c | c |B>{\columncolor{highlight1}} c | c | c | c |}
            \hline \textbf{Year} & \textbf{Country} & \textbf{\color{frame}
COUNT(\color{black}City\color{frame})} & ...\\ \hline
			1896 & Greece & \cellcolor{highlight2}\colorboxed{frame}{Athens} & ...\\ \hline
			1900 & France & Paris & ...\\ \hline
			... & ... & ... & ...\\ \hline
			2004 & Greece & \cellcolor{highlight2}\colorboxed{frame}{Athens} & ...\\ \hline
			2008 & China & Beijing & ...\\ \hline
			2012 & UK & London & ...\\ \hline
            2016 & Brazil & Rio de Janeiro & ...\\ \hline
        \end{tabular}
        \caption{Aggregation}\label{tbl:highlight_aggregation}
        \small
        \textit{the number of rows where value of column City is Athens.\newline}
    \end{minipage}%
    \begin{minipage}{.6\linewidth}
        \centering
        \begin{tabular}{|c |B>{\columncolor{highlight1}} c | c | c | c |B>{\columncolor{highlight1}}c |}
            \hline \textbf{Rank} & \textbf{Nation} & \textbf{Gold} & \textbf{Silver} & \textbf{Bronze} & \textbf{Total} \\ \hline
			1 & New Caledonia & 120 & 107 & 61 & 288\\ \hline
            2 & Tahiti & 60 & 42 & 42 & 144\\ \hline
            3 & Papua New Guinea & 48 & 25 & 48 & 121\\ \hline
            4 & \colorboxed{frame}{Fiji} & 33 & 44 & 53 & \cellcolor{highlight2}\colorboxed{frame}{130} \\ \hline
            5 & Samoa & 22 & 17 & 34 & 73\\ \hline
            6 & Nauru & 8 & 10 & 10 & 28\\ \hline
            7 & \colorboxed{frame}{Tonga} & 4 & 6 & 10 & \cellcolor{highlight2}\colorboxed{frame}{20} \\ \hline
            ... & ... & ... & ... & ... & ...\\ \hline
    %\hspace*{-0.4cm}
        \end{tabular}
        \caption{Difference (values)}\label{tbl:highlight_diff_val}
        \small
        \textit{difference in column Total between Fiji and Tonga.\newline}
    \end{minipage}

    \begin{minipage}{.5\linewidth}
        \centering
        \begin{tabular}{|c |B>{\columncolor{highlight1}} c | c | c | c | c |}
            \hline \textbf{Temple} & \textbf{Town} & \textbf{Prefecture} & ...\\ \hline
			Iwaya-ji & Kumakogen & Ehime Prefecture & ...\\ \hline
            Yakushi Nyorai & \cellcolor{diff1}\colorboxed{frame}{Matsuyama} & Ehime Prefecture & ...\\ \hline
            Amida Nyorai & \cellcolor{diff1}\colorboxed{frame}{Matsuyama} & Ehime Prefecture & ...\\ \hline
            Shaka Nyorai & \cellcolor{diff1}\colorboxed{frame}{Matsuyama} & Ehime Prefecture & ...\\ \hline
            Yakushi Nyorai & \cellcolor{diff1}\colorboxed{frame}{Matsuyama} & Ehime Prefecture & ...\\ \hline
            Yokomine-ji & Saijo & Ehime Prefecture & ...\\ \hline
            Fudo Myoo & \cellcolor{diff2}\colorboxed{frame}{Imabari} & Ehime Prefecture & ...\\ \hline
            Jizo Bosatsu & \cellcolor{diff2}\colorboxed{frame}{Imabari} & Ehime Prefecture & ...\\ \hline
            ... & ... & ... & ...\\ \hline
        \end{tabular}
        \caption{Difference (occurrences)}\label{tbl:highlight_diff_occur}
        \small
        \textit{in column Town, what is the difference between rows with value Matsuyama and rows with value Imabari.\newline}
    \end{minipage}%
    \begin{minipage}{.33\linewidth}
        \centering
        \begin{tabular}{|c |B>{\columncolor{highlight1}} c |B>{\columncolor{highlight1}} c | c | c | c |}
            \hline \textbf{Year} & \textbf{Country} & \textbf{City} & ...\\ \hline
			1896 & \colorboxed{frame}{Greece} & \cellcolor{highlight2}\colorboxed{frame}{Athens} & ...\\ \hline
			1900 & France & Paris & ...\\ \hline
			... & ... & ... & ...\\ \hline
			2004 & \colorboxed{frame}{Greece} & \cellcolor{highlight2}\colorboxed{frame}{Athens} & ...\\ \hline
			2008 & \colorboxed{frame}{China} & \cellcolor{highlight2}\colorboxed{frame}{Beijing} & ...\\ \hline
			2012 & UK & London & ...\\ \hline
            2016 & Brazil & Rio de Janeiro & ...\\ \hline
    %\hspace*{-0.4cm}
        \end{tabular}
        \caption{Union}\label{tbl:highlight_or}
        \small
        \textit{values of column City where value of column Country is China or Greece.\newline}
    \end{minipage}
    \begin{minipage}{.33\linewidth}
        \centering
        \begin{tabular}{|B>{\columncolor{highlight1}} c |B>{\columncolor{highlight1}} c |B>{\columncolor{highlight1}} c | c | c | c |}
            \hline \textbf{Year} & \textbf{Country} & \textbf{City} & ...\\ \hline
			1896 & Greece & Athens & ...\\ \hline
			1900 & France & Paris & ...\\ \hline
			... & ... & ... & ...\\ \hline
			2004 & Greece & Athens & ...\\ \hline
			2008 & China & Beijing & ...\\ \hline
			\colorboxed{frame}{2012} & \colorboxed{frame}{UK} & \cellcolor{highlight2}\colorboxed{frame}{London} & ...\\ \hline
            2016 & Brazil & Rio de Janeiro & ...\\ \hline
    %\hspace*{-0.4cm}
        \end{tabular}
        \caption{Intersection}\label{tbl:highlight_and}
        \small
        \textit{values of column City where value of column Country is UK and also where value of column Year is 2012.\newline}
    \end{minipage}
    \begin{minipage}{.33\linewidth}
        \centering
        \begin{tabular}{|B>{\columncolor{highlight1}} c | c |B>{\columncolor{highlight1}} c | c | c | c |}
            \hline \textbf{Year} & \textbf{Country} & \textbf{City} & ...\\ \hline
			1896 & Greece & Athens & ...\\ \hline
			1900 & France & Paris & ...\\ \hline
			... & ... & ... & ...\\ \hline
			2004 & Greece & Athens & ...\\ \hline
			\colorboxed{frame}{2008} & China & \colorboxed{frame}{Beijing} & ...\\ \hline
			\cellcolor{highlight2}\colorboxed{frame}{2012} & UK & \colorboxed{frame}{London} & ...\\ \hline
            2016 & Brazil & Rio de Janeiro & ...\\ \hline
    %\hspace*{-0.4cm}
        \end{tabular}
        \caption{Superlative (values)}\label{tbl:highlight_superlative_val}
        \small
        \textit{between London or Beijing who has the highest value of column Year.\newline}
    \end{minipage}
    \begin{minipage}{.33\linewidth}
        \centering
        \begin{tabular}{| c | c |B>{\columncolor{highlight1}} c | c | c | c |}
            \hline \textbf{Year} & \textbf{Country} & \textbf{City} & ...\\ \hline
			1896 & Greece & \cellcolor{highlight2}\colorboxed{frame}{Athens} & ...\\ \hline
			1900 & France & \colorboxed{frame}{Paris} & ...\\ \hline
			... & ... & ... & ...\\ \hline
			2004 & Greece & \cellcolor{highlight2}\colorboxed{frame}{Athens} & ...\\ \hline
			2008 & China & \colorboxed{frame}{Beijing} & ...\\ \hline
			2012 & UK & \colorboxed{frame}{London} & ...\\ \hline
            2016 & Brazil & Rio de Janeiro & ...\\ \hline
    %\hspace*{-0.4cm}
        \end{tabular}
        \caption{Superlative (occurrences)}\label{tbl:highlight_superlative_occur}
        \small
        \textit{the value that appears the most in column City.\newline}
    \end{minipage}

\end{table*}

\newpage

\bibliographystyle{abbrv}
\bibliography{vldb_sample}  % vldb_sample.bib is the name of the Bibliography in this case

% References are generated by bibtex from your ~.bib file.  Run latex,
%then bibtex, then latex twice (to resolve references).

%APPENDIX is optional.
% ****************** APPENDIX **************************************
% Example of an appendix; typically would start on a new page
%pagebreak
%\begin{appendix}
%\end{appendix}

\end{document}